\documentclass[sigconf,natbib=false]{acmart}
\AtBeginDocument{%
  }

\copyrightyear{2023}
\acmYear{2023}
\setcopyright{acmlicensed}
\acmConference[CCS '23]{Proceedings of the 2023
ACM SIGSAC Conference on Computer and Communications
Security}{November 26--30, 2023}{Copenhagen, Denmark}
\acmBooktitle{Proceedings of the 2023 ACM SIGSAC Conference on Computer
and Communications Security (CCS '23), November 26--30, 2023, Copenhagen,
Denmark}
\acmPrice{15.00}
\acmDOI{10.1145/3576915.3616652}
\acmISBN{979-8-4007-0050-7/23/11}

\RequirePackage[
  datamodel=acmdatamodel,
  style=acmnumeric,
  ]{biblatex}

\usepackage{comment}
\usepackage{makecell}
\usepackage{multirow}
\newcommand{\paragraphb}[1]{\noindent{\bf #1} }
\newcommand{\paragraphe}[1]{\vspace{0.03in} \noindent{\em #1} }

\begin{document}

\fancyfoot[C]{\thepage}
\title{Stealing the  Decoding Algorithms of  Language Models}

\author{Ali Naseh}
\affiliation{%
  \institution{University of Massachusetts Amherst}
  \city{Amherst}
  \state{Massachusetts}
  \country{USA}
}
\email{anaseh@cs.umass.edu}

\author{Kalpesh Krishna}
\affiliation{%
  \institution{University of Massachusetts Amherst}
  \city{Amherst}
  \state{Massachusetts}
  \country{USA}
}
\email{kalpesh@cs.umass.edu}

\author{Mohit Iyyer}
\affiliation{%
  \institution{University of Massachusetts Amherst}
  \city{Amherst}
  \state{Massachusetts}
  \country{USA}
}
\email{miyyer@cs.umass.edu}

\author{Amir Houmansadr}
\affiliation{%
  \institution{University of Massachusetts Amherst}
  \city{Amherst}
  \state{Massachusetts}
  \country{USA}
}
\email{amir@cs.umass.edu}

\newcommand{\grumbler}[2]{\textcolor{red}{\bf #1: #2}}
\newcommand{\mgrumbler}[2]{\textcolor{cyan}{\bf #1: #2}}
\newcommand{\ogrumbler}[2]{\textcolor{orange}{\bf #1: #2}}
\newcommand{\ggrumbler}[2]{\textcolor{violet}{\bf #1: #2}}
\newcommand{\red}[1]{\textcolor{red}{#1}}
\newcommand{\better}[1]{(\textcolor{blue}{$-#1$})}
\newcommand{\worse}[1]{(\textcolor{red}{$+#1$})}

\newcommand{\fade}[1]{\textcolor{gray}{#1}}

\newcommand{\milad}[1]{\mgrumbler{Milad}{#1}}
\newcommand{\ali}[1]{\ggrumbler{Ali}{#1}}
\newcommand{\amir}[1]{\grumbler{amir}{#1}}

\newcommand{\micomment}[1]{\textcolor{brown}{\bf \small [ #1 --MI]}}
\newcommand{\kkcomment}[1]{\textcolor{blue}{\bf \small [ #1 --KK]}}

\begin{abstract}
  A key component of generating text from modern language models (LM) is the selection and tuning of \emph{decoding algorithms}. These algorithms determine how to generate text from the internal probability distribution generated by the LM. The process of choosing a decoding algorithm and tuning its hyperparameters takes significant time, manual effort, and computation, and it also requires extensive human evaluation. Therefore, the identity and hyperparameters of such decoding algorithms are considered to be extremely valuable to their owners. In this work, we show, for the first time, that an adversary with typical API access to an LM can steal the type and hyperparameters of its decoding algorithms at very low monetary costs. Our attack is effective against popular LMs used in text generation APIs, including GPT-2, GPT-3 and GPT-Neo. We demonstrate the feasibility of stealing such information with only a \emph{few dollars}, e.g., $\$0.8$, $\$1$, $\$4$, and $\$40$ for the four versions of GPT-3.

\end{abstract}



\maketitle

\section{Introduction}

Language models (LM) have become a crucial part of various text generation APIs, such as machine translation, question answering, story generation, and text summarization. Large-scale LMs like GPT-2~\cite{radford2019language}, GPT-3~\cite{brown2020language} and GPT-Neo~\cite{black2022gpt} have been shown to generate high-quality texts for these tasks. To generate a sequence of tokens, LMs produce a probability distribution over the vocabulary at each time step, from which the predicted token is drawn. Enumerating all possible output sequences for a given input and choosing the one with the highest probability is intractable; furthermore, relatively low-probability sequences may even be desirable for certain tasks (e.g., creative writing). Therefore, LMs rely on \emph{decoding algorithms} to decide which output tokens to produce based on their probabilities, i.e., to decode the text.  

As shown in the literature~\cite{dou2021scarecrow}, the choice of the decoding algorithm and its hyperparameters is critical to the performance of the LM on text generation tasks. Thus, users of many LM-based APIs are offered a choice of decoding algorithms and also the ability to adjust any corresponding hyperparameters. For example, in machine translation, beam search is more common than other methods; however, in story generation, sampling-based methods are preferred for their ability to generate more diverse text~\cite{see-etal-2019-massively}.

\begin{figure*}[h]
 \centering
 \includegraphics[width = 1.0\linewidth]{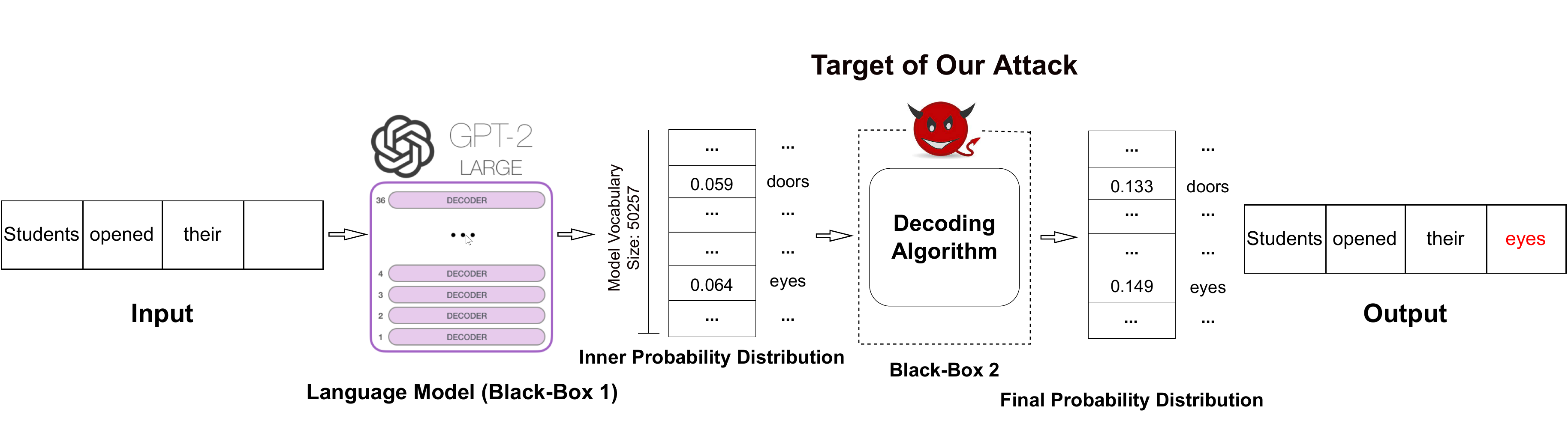}
 \caption{Overview of a typical  LM-based API; it includes two independent black boxes: an LM which generates a probability distribution over the vocabulary, and a  decoding algorithm which dictates text generation based on the LM's inner probability distribution. 
 }
 \vspace{-2ex}
 \label{fig:training}
\end{figure*}


Deciding and fine-tuning the decoding algorithm is a costly operation  in typical text generation tasks. This is because automatic metrics poorly reflect quality, so human evaluation is needed  to tune the decoding algorithms~\cite{gehrmann2022repairing, celikyilmaz2020evaluation}. 
That is,  the service provider needs to perform a manual human evaluation to find the best decoding algorithm and the corresponding hyperparameter(s). For instance, they need to recruit people to read and evaluate the generated texts manually, a process that  could be costly. There is also the cost of developing and maintaining the evaluation infrastructure, such as software and hardware used to conduct the evaluations, and the cost of analyzing and interpreting the results. 
To summarize, \textbf{decoding algorithms are considered to be significant assets  of conventional LM systems}. 
We refer the reader to Section ~\ref{sec:motivation} for a more elaborate discussion on the value of decoding algorithms with an example scenario. 


In this paper, we ask the following question: \textbf{Can an adversary steal (i.e., infer) the type of decoding algorithm in an LM-based systems, as well as its corresponding hyperparameter(s) by merely accessing the text generation APIs? And if yes, at what  monetary costs?}
We show that it is indeed possible to infer the type and hyperparameters of a deployed decoding algorithm \emph{with high accuracies and at low costs}! 

To the best of our knowledge, our paper is the first to explore  decoding algorithm  stealing attacks on  LM systems. 
While there exist model stealing attacks in other machine learning (ML) contexts (like vision tasks~\cite{orekondy2019knockoff}), such stealing techniques can not be applied to the setting of LMs. 
Unlike computer vision and text classification tasks, text generation systems are composed of two cascaded building black boxes,  as shown in Figure~\ref{fig:training}: the language model and the decoding algorithm. In this setting, the adversary aims to attack the second black box, given some (public) knowledge about the first block. Therefore, this is a unique problem like no other ML stealing attack, which requires tailored attack algorithms.

\paragraphb{Overview of our stealing attack:}
The main \emph{intuition} of our attack is  that \emph{different decoding algorithms and different values of  hyperparameters can leave distinguishable signatures on the text being generated by LM-based APIs~\cite{dou2021scarecrow}.}
For instance, Table ~\ref{tab:Decoding_Examples2} 
shows how different decoding algorithms can lead to entirely different API outputs  given the exact same input text.  
We, therefore, obtain analytical algorithms that aim to infer the type and hyperparameters of the decoding algorithms based on their API observations. 

Our attack aims to distinguish between the widely used decoding algorithms, i.e., greedy search, beam search, pure sampling, sampling with temperature, 
top-k sampling,  Nucleus Sampling, and even the combination of these algorithms, just by observing the output of the LM APIs. Notably, these six decoding algorithms are currently the most widely utilized. While several alternative algorithms have been proposed recently (e.g., RankGen decoding~\cite{krishna2022rankgen}), none have been implemented in any publicly available LM-based API. Moreover, our approach can be adapted to accommodate new decoding algorithms. To provide a tangible example, the most popular LM API (OpenAI's GPT-3) offers users a choice of only three decoding algorithms, all of which can be detected by our method.

Specifically, we design a multi-stage attack algorithm that leverages the text generation API.
We  assume that the adversary has access to the probability distributions provided by the target LM, which is 
 a standard assumption in 
 the NLP literature~\cite{shokri2017membership,yang2019adversarial, wang2018stealing, carlini2021extracting} and valid in the real world, e.g., the GPT-3 API gives probability distribution for top tokens, and many causal LMs are open-source~\cite{radford2019language,zhang2022opt,nijkamp2022conversational}. In Section ~\ref{sec:pre-requisites}, we demonstrate three scenarios in which the attacker can obtain the necessary information, namely the inner probability distribution provided by the API's LM. Furthermore, in Section ~\ref{sec:other_discussion} and Appendix ~\ref{sec:other_discussion2}, we demonstrate how the attacker can apply our proposed attack in each of these scenarios and the extent of information that can be stolen.
%

%


\paragraphb{Summary of our results:} We assessed the efficacy of our proposed attack by testing it on three of the most prominent LMs: GPT-2, GPT-3, and GPT-Neo. We conducted experiments using various sizes of GPT-2 and GPT-3 
to demonstrate that the size of the model does not affect the results of our attack. In other words, our attack demonstrates consistent performance across different sizes of these models. We achieve almost perfect accuracy in detecting the type of decoding algorithm while also obtaining accurate estimates of the hyperparameters. All of these can be done with a few dollars, e.g., $\$0.8$, $\$1$, $\$4$, and $\$40$ for the four versions of GPT-3 (more details in Section~\ref{sec:Cost}). All our experiments are centered around employing a custom decoding algorithm in collaboration with GPT-2/3/Neo models, as opposed to targeting a real deployment of these LMs.


In all cases, the probability distribution resulting from the estimated decoding algorithms was found to be similar to the probability distribution provided by the targeted API. The high p-value and low KL-divergence score in all cases confirm our claims. Furthermore, we provide both theoretical and practical evaluations of the number of queries required to perform accurate attacks.

Finally, we introduce a potential defense against our attack and assess its implications on the quality of the generated text. Our evaluations show the effectiveness of our countermeasure, i.e., the inferred hyperparameters significantly deviate  from their true values in the presence of our countermeasure. For instance, upon deploying our defense mechanism, we observe that the inferred values for the temperature and $p$ in Nucleus Sampling will, on average, differ by 0.1 from their actual values, representing a notable discrepancy. Given that these hyperparameters typically fall within the range of 0 to 1, such deviations can lead to varying behaviors in LM systems. On the other hand, our countermeasure does not substantially affect the utility of the system as measured by the perplexity feature.


\section{Background}

In this section, we will review some basic concepts from language modeling. Many architectures have been proposed to be used as LMs in various applications. Graves~\cite{graves2013generating} introduces Long Short-Term Memory recurrent neural networks to generate real-valued sequences with long-range structures to predict the next token. Bahdanau et al.~\cite{bahdanau2014neural} present an encoding-decoding approach using an attention mechanism for machine translation systems. More recently, Vaswani et al.~\cite{46201} introduce an attention-based mechanism, called Transformers, to replace the former RNN-based models. All the large-scale NLP algorithms are using transformers, like BERT~\cite{devlin2018bert}, T5~\cite{raffel2020exploring}, GPT-2~\cite{radford2019language}, GPT-3~\cite{brown2020language}. We will skip the details of the architectures and refer the reader to the mentioned works. 

\subsection{Open-Ended Text Generation} Many text generation tasks use autoregressive LMs to generate text. For some of these tasks, including machine translation~\cite{bahdanau2014neural} and summarization~\cite{nallapati2016abstractive, zhong2020extractive}, the output is more constrained due to the input. However, diverse outputs are desirable in several NLP tasks, including story generation and Question-Answering. As described in~\cite{clark2018neural, holtzman2018learning}, the task of open-ended  text generation  is to generate text that forms a coherent continuation from the given context. In other words, suppose that $w_1, w_2,..., w_l$ are tokens of a sequence from a vocabulary \textit{V}; we want the model to generate $r$ continuation tokens in a left-to-right fashion by applying the chain rule of probability: 

\begin{equation}
    Pr(w_{1:l+r}) = \prod_{i=1}^{l+r}Pr(w_i|w_{1:i-1})
\end{equation}

The next step after computing the conditional probabilities, we need to decide how we want to pick the next token based on the probabilities.

\subsection{Decoding Algorithms}

In this section, we overview the six most commonly used decoding algorithms. Two of these approaches are deterministic (greedy decoding and beam search), and the other four use probabilistic sampling techniques (random sampling, sampling with temperature, top-k sampling, and Nucleus Sampling). Examples of texts generated by various decoding algorithms can be found in Table ~\ref{tab:Decoding_Examples2}.

\subsubsection{Maximization-Based Methods}

We assume that the model assigns higher probability scores to higher quality text in maximization-based decoding approaches~\cite{holtzman2019curious}. Hence, these methods search for continuation to maximize the probability of the generated sequence. Greedy and Beam search are two prominent maximization-based decoding methods.

As a simple method, \textbf{Greedy Search} selects the word with the highest probability as the next token: $w_i = \arg\max_{w} Pr(w | w_{1:i-1})$ at each time step $t$. However, this approach may result in locally optimal, but globally sub-optimal decisions, where high probability paths are not encountered. To mitigate this limitation, \textbf{Beam Search} generates a specified number of hypotheses at each time step, and selects the token sequence with the highest probability among them. However, research has shown that text generated by both of these methods can lack quality and diversity, even with a high number of beams, as reported in studies such as~\cite{45936, vijayakumar2016diverse}. Although beam search is effective in tasks with constrained outputs such as machine translation, it fails to perform well in open-ended text generation scenarios.

\begin{table*}[ht]
    \centering
    \caption{\label{tab:Decoding_Examples2} These are some text examples generated by the API using various decoding algorithms. The initial prompt is "Yesterday, I have decided to"}
    \vspace{-2ex}

    \label{tab:Decoding_Examples2}

    \resizebox{1.0\textwidth}{!}{
    \begin{tabular}{cl}
    \toprule
        Decoding Algorithm & Generated Text \\
        \midrule
        Greedy Search &  Yesterday, I have decided to write a blog post about the recent events in the United States. I will be writing about the events \\ 
        Beam Search &   Yesterday, I have decided to take a look at some of the more interesting things that have happened in the last couple of weeks. \\
        Pure Sampling &  Yesterday, I have decided to go to Africa for the spiritual transformation, and to have a hard time making it work if I don \\
        Sampling with Temperature & Yesterday, I have decided to start my 6th season as a fan. My goal is always to make my class stand again. \\
        Top-k Sampling & Yesterday, I have decided to go ahead and continue contributing to the discussion here on RSI (RSS Feed). In short —\\
        Nucleus Sampling &  Yesterday, I have decided to write something more in French! It's something that I was at the Festival last year, in Canada\\
        Top-k and Nucleus Sampling &  Yesterday, I have decided to post this post. I was planning to post it a few days before the deadline, but have decided \\
        Temperature, Top-k and Nucleus Sampling &  Yesterday, I have decided to make a post for your consideration. Please note that I am not a psychologist.\\
        \bottomrule
        \end{tabular}
        }
    \vspace{-2mm}
\end{table*}

\subsubsection{Sampling-Based Methods}


In open-ended text generation, to have a more diverse output, it is suggested to use probabilistic (i.e., sampling-based) decoding algorithms~\cite{see-etal-2019-massively}. The basic sampling method is the simple \textbf{random sampling} using conditional probability distribution at each time step. However, the major drawback of this approach is that any irrelevant token has a chance to be picked~\cite{holtzman2019curious}, so it can lead to an inappropriate and irrelevant output. Below we introduce alternative probabilistic decoding mechanisms that are commonly used.

\textbf{Top-K Sampling.} In this approach~\cite{fan2018hierarchical}, at each time step, the algorithm picks $k$ tokens with the highest probabilities from the distribution over the vocabulary. Then, we re-scale the probabilities of these $k$ tokens to sum $1$. Now, we can sample from resulted $k$ tokens. In other words, with given probability distribution $Pr(w|w_{1:i-1})$ and the set of $k$ tokens with the highest probabilities $V^{(k)}$, we will sample from the new distribution:

\begin{equation}
    Pr'(w|w_{1:i-1}) = \begin{cases} 
      \frac{Pr(w|w_{1:i-1})}{S} & if w \in V^{(k)} \\
      0 & otherwise
   \end{cases}
\end{equation}

\noindent where $ S = \sum_{w \in V^{(k)}} Pr(w|w_{1:i-1})$. We can create more human-like text using the Top-k approach compared to the basic sampling. However, choosing an appropriate $k$ for a specific task is always a concern. Also, after setting $k$, we will use the same $k$ for all time steps, which is problematic. In some time steps, the probability distribution might be flat, and in some other time steps might be peaked. So, a fixed $k$ may not be a good choice for some time steps. Nucleus Sampling aims to address  this issue.

\textbf{Nucleus Sampling.} Unlike top-k sampling, the Nucleus Sampling~\cite{holtzman2019curious}  algorithm does not pick a fixed number of tokens. Instead, it picks the smallest number of tokens whose cumulative probability exceeds $p$. Then, it re-scales these probabilities to sum $1$. Then, we can sample from the new distribution. Similarly, suppose that we are given probability distribution $Pr(w|w_{1:i-1})$ and the smallest set of tokens $V^{(p)}$ whose cumulative probability exceeds $p$, $\sum_{w \in V^{(p)}} Pr(w|w_{1:i-1}) \geq p$. We will sample from the new distribution:

\begin{equation}
    Pr'(w|w_{1:i-1}) = \begin{cases} 
      \frac{Pr(w|w_{1:i-1})}{S} & if w \in V^{(p)} \\
      0 & otherwise
   \end{cases}
\end{equation}

\noindent where $ S = \sum_{w \in V^{(p)}} Pr(w|w_{1:i-1})$. This dynamic selection approach leads to generate more human-like texts.

\textbf{Sampling with Temperature.}
Another popular alternative to the basic random sampling technique is adding temperature to the probability distribution~\cite{ackley1985learning}. This approach has been used in various text generation applications~\cite{ficler2017controlling, DBLP:conf/iclr/CacciaCFLPC20}. As described before, in the basic random sampling method, any token even with low conditional probability has the chance to be picked. If we apply temperature to the softmax, we will amplify the likelihood of high probable tokens and attenuate the likelihood of low probable ones. More formally, suppose that $u_{1:|V|}$ are our logits, and we are given the temperature $t$. Then, the new softmax formula will be:

\begin{equation}
    Pr'(w=V_j|w_{1:i-1}) = \frac{\exp (\frac{u_j}{t})}{\sum_{k=1}^{|V|} \exp(\frac{u_k}{t})} 
\end{equation}

\noindent 
Recent works show that using smaller  $t$ decreases the diversity even though the quality of generated text increases~\cite{hashimoto2019unifying}.

\section{High-level Overview of Our Attack}\label{sec:Stealing Attack}

In this section, we first establish the key terminology used throughout the paper. We then go through the adversary's capabilities, objectives, and targets. Following that, we provide a detailed example to motivate the problem. Finally, we elaborate on different scenarios in which the adversary can acquire the inner probabilities.

\subsection{Terminologies}
We define two types of probability distributions which will be referenced throughout our algorithms. The \emph{inner probability distribution} represents the probability distribution generated by the LM before being fed into the decoding section. The \emph{final probability distribution} represents the probability distribution of tokens generated by the API as a whole after the application of the decoding algorithm. These two probability distributions are illustrated in Figure ~\ref{fig:training}.

\subsection{Threat Model} 

\paragraphb{Adversary's Capabilities.} \label{sec:Capabilities}As previously discussed, open-ended text generation APIs consist of two independent components: a \emph{language model} and a \emph{decoding algorithm}. In our attack, the adversary has black-box access to both of these components, meaning that the attacker has no information about the specifics of the decoding algorithm. Additionally, in the majority of cases, text generation APIs operate using a restricted range of models, including GPT-3. These models exhibit distinct behaviors that can lead to different orders of tokens when sorted by their probabilities at a given time step; decoding algorithms then only adjust the weighting assigned to each token. This characteristic facilitates the identification of the base model used by the API through output comparison. Hence, the adversary does not need to know the details of the targeted LM. The only information the attacker has is the inner probability distribution. \emph{Specifically, the attacker only needs the probabilities of the top two tokens to apply all stages of the attack}. In Section ~\ref{sec:pre-requisites}, we describe in detail how the attacker can get such information.

\paragraphb{Adversary's Objective.} In our attack, the first step is for the adversary to infer the type of decoding algorithm used by the API and subsequently extract any corresponding hyperparameters. The ultimate goal is that by using the extracted decoding type and corresponding hyperparameters, the final probability distribution of generated tokens should be the same as that of the victim model.

\paragraphb{Attack Target.} In this study, we selected GPT-2~\cite{radford2019language}, GPT-3~\cite{brown2020language}, and GPT-Neo~\cite{black2022gpt} as the victim models. GPT-2 is a large transformer-based LM that was pre-trained on a dataset of 8 million web pages. GPT-3 is another large LM that was pre-trained on a dataset of terabytes of text data, comprising a diverse range of web pages, books, articles, and other text sources. GPT-Neo is an open-source, large-scale language model developed by EleutherAI. GPT-Neo is designed to provide similar capabilities and performance as OpenAI's GPT-3 while being accessible to the broader research community. GPT-Neo has been pre-trained on the Pile~\cite{gao2020sid}, a diverse dataset comprising over 800 gigabytes of text data developed by EleutherAI. All of these models have outstanding capabilities in text generation tasks. Since GPT-2 is available free of charge and we needed to send millions of queries for our experiments, it was more feasible to conduct the majority of our experiments using this model.

\subsection{Attack's Motivation} \label{sec:motivation}

Many large-scale LMs developers, including OpenAI, Google, Amazon, and Microsoft, have trained their own LMs and released them via API access for various tasks. Concurrently, numerous companies, primarily startups such as Jasper\footnote{https://www.jasper.ai} and perplexity.ai\footnote{https://www.perplexity.ai}, have launched products that essentially serve as wrappers around these APIs. Tuned decoding algorithms constitute a significant competitive advantage for these companies, as considerable time and effort are required for optimization. Our paper demonstrates that any party with access to their own LM (i.e., any of these companies) can employ low-cost attacks to pilfer decoding algorithms associated with other APIs, posing a concern for businesses that depend on LMs.

To further elucidate the significance of this issue, we present a pertinent example. We reference the data provided in ~\cite{dou2021scarecrow}, which detail the costs associated with their annotation process using Amazon Mechanical Turk. Suppose a company seeks to identify the optimal decoding algorithm and corresponding hyperparameters from nine different configurations, comprising three decoding algorithms and three distinct hyperparameters for each. The company generates 100 paragraphs using each configuration and enlists 20 crowd workers to assess them. Based on the data in ~\cite{dou2021scarecrow}, if the company pays $\$3.5$ per paragraph, the total cost for evaluating all configurations amounts to $\$63,000$. Additionally, there are expenses related to designing and administering qualification exams for these crowd workers. These numbers underscore the importance of safeguarding such information.

\subsection{Attack's Prerequisites} \label{sec:pre-requisites}

As previously discussed, the attacker only requires knowledge of the inner probability distribution generated by the API's LM, specifically the top two tokens. There are three scenarios in which an attacker can acquire this prerequisite information.

First, the attacker may have direct access to the probability distribution of the LM, as some APIs such as OpenAI GPT-3 provide the probabilities of the top tokens, which are sufficient for our attack. 

Second, the attacker may not have direct access to the probability distribution but can approximate these values. For example, if the APIs use unmodified base models, the attacker can use the base model as a reference. Additionally, in some LM-based tools and APIs that use \emph{prompt engineering} instead of fine-tuning, the attacker can still use the base model as an approximation of the inner probability distribution. There are various platforms that allow users to generate high-quality written content such as blog posts, product descriptions, and other personalized content. Such APIs usually rely on prompt engineering instead of fine-tuning these models. In Section ~\ref{sec:other_discussion}, we show that in such cases the attacker can still use the base model as the reference model.

Third, the attacker can employ a model-stealing approach to infer these probabilities. It is important to note that model stealing attacks are a parallel research direction that aims to extract the inner probability distribution by targeting the first black box of the text generation system. However, our focus is on attacking the second black box, the decoding algorithm. Using a model-stealing approach is typically employed when the API fine-tunes a pre-trained model~\cite{49091}. However, it is essential to consider that the fine-tuning process can alter the weights of the model.

\subsection{High-level Attack Approach}

Our mathematical-based attack leverages the properties of each decoding method to detect them in consecutive stages (as shown in Figure ~\ref{fig:algorithm}). For example, we utilize statistical properties and the probability distribution generated by these decoding approaches in sampling-based methods. In addition to detecting the type, we will also aim to estimate the hyperparameters. To achieve this, we provide mathematical formulas that utilize the internal probabilities provided by the API's LM, as well as the final probabilities generated by the API. We also demonstrate the theoretical reasoning behind why these formulae result in accurate estimation.

\section{Details of Our Attack Algorithms}\label{subsec:attack}

In this section, we propose multi-stage algorithm to extract the type and then the corresponding hyperparameter(s) used in the API. To make all of our experiments and analysis more consistent, we will focus on the task of text completion. 

\paragraphb{Stage 1: Is it a sampling decoding algorithm or not?}
In the first step, we aim to determine whether the decoding algorithm is a sampling-based method or not. With sampling methods, the output vary if the same query is sent to the API multiple times. However, if the API employs a non-sampling method and an arbitrary prompt is sent to the API, the output will be the same. In this case, we can conclude that the decoding algorithm is either greedy or beam search. However, it's possible that for very small values of $p$ or $k$ (although this is unlikely to occur in practical applications), the same behavior is observed. In such cases, increasing the number of queries reduces the probability of this outcome to close to zero.

\paragraphb{Stage 2: Is it greedy or beam search?} In the second step, we aim to determine whether the decoding algorithm is greedy or beam search. In greedy search, at each time step, the most probable token is selected from the probability distribution provided by the model. In other words, when starting with an arbitrary sequence and generating tokens one at a time, at each time step $t$, the subsequences $s_{1}s_{2}...s_{i}$ for $i<t$ will not change. An example of this can be seen in Table ~\ref{tab:Decoding_Examples}.

However, this is not necessarily the case in beam search. Suppose that at time step $t$, the token $s_t$ is the most probable token from the probability distribution provided by the model. If we query the sequence $s_{1}s_{2}...s_{t}$ to the API to generate the sequence $s_{1}s_{2}...s_{t+1}$, in the generated sequence, $s_t$ is not necessarily the same as the $s_t$ in the previous time step. Hence, if we start with an arbitrary sequence and try to complete it token by token and if the subsequences in the output do not change, it means that with a high probability, it is greedy; otherwise, it is beam search.

An example is provided in Table~\ref{tab:Decoding_Examples}. In this example, the API generates "to" at time step $t+1$ and index $i$. After generating the token at time step $t+4$, we do not see "to" in the subsequence at the index $i$ anymore, and this is due to the nature of beam search.
To increase this probability, we can repeat this experiment for more time steps and more arbitrary sequences. We will discuss the required number of queries in Section ~\ref{sec:exp}. In Table ~\ref{tab:Decoding_Examples}, some examples of generated text using these two decoding strategies for consecutive tokens are presented.

After detecting beam search as the decoding algorithm, we must estimate the beam size as its hyperparameter. Again, suppose that we start with the arbitrary sequence $s_1, s_2, ..., s_{t-1}$, and we plan to generate tokens one by one. In the first step, we generate the token $s_t$. Then, we use the sequence $s_1, s_2, ..., s_t$ and try to generate the token at time step $t+1$. If we continue with the same process, the token at position $t$ may not be the same as the token we generated in the first step. After repeating this process multiple times, we might have different tokens generated at the position $t$. Then, we find the rank of these tokens from the sorted tokens based on their probabilities provided by the model. The maximum rank among these tokens is our estimation of the beam size. To make our estimation more reliable, we can repeat this process for multiple arbitrary sequences and pick the maximum value as our final estimation. In Appendix ~\ref{sec:beam_size}, we provide more detailed examples to further illustrate this method.

\begin{table}[t]
    \caption{\label{tab:Decoding_Examples} This table presents some text examples generated by the API using the greedy or beam search decoding approach for consecutive time steps. The examples are generated using the initial prompt "Students opened their". It showcases the difference in the generated text by these two decoding approach.}
    \vspace{-2ex}
    \label{tab:Decoding_Examples}
    \resizebox{0.48\textwidth}{!}{
    \begin{tabular}{clc}
    \toprule
        Time Step & Generated Text  & Approach\\
        \midrule
        t & Students opened their doors  & Greedy Search \\
        t+1 & Students opened their doors \textcolor{red}{to} & Greedy Search \\
        t+2 & Students opened their doors \textcolor{red}{to} the & Greedy Search \\
        t+3 & Students opened their doors \textcolor{red}{to} the public & Greedy Search \\
        t+4 & Students opened their doors \textcolor{red}{to} the public on & Greedy Search \\
        t+5 & Students opened their doors \textcolor{red}{to} the public on Friday & Greedy Search \\
        \midrule
        t & Students opened their doors & Beam Search \\
        t+1 & Students opened their doors \textcolor{red}{to} & Beam Search \\
        t+2 & Students opened their doors \textcolor{red}{to} the & Beam Search \\
        t+3 & Students opened their doors \textcolor{red}{to} the public & Beam Search \\
        t+4 & Students opened their doors \textcolor{red}{for} the first time & Beam Search \\
        t+5 & Students opened their doors \textcolor{red}{at} 6 p.m & Beam Search \\
        \bottomrule
        \end{tabular}
        }
    \vspace{-3mm}
\end{table}

\paragraphb{Detecting combined decoding strategies:} In the sampling-based decoding algorithms, note that in some cases, the decoding algorithm can be a combination of more than one decoding algorithm. In particular, the following cases are common:

  1)  only temperature;
  2)  only top-k sampling;
  3)  only Nucleus Sampling;
  4)  only random sampling;
  5)  both temperature and top-k Sampling;
  6)  both temperature and Nucleus Sampling;
  7)  both top-k and Nucleus Sampling;
  8)  all temperature, top-k, and Nucleus Sampling.

In other words, we want to determine which combination of decoding algorithms the API uses, and then determine the corresponding hyperparameter(s). In all cases, we send an arbitrary sequence to the API multiple times. Then, we sort all tokens generated at time step $t$ in descending order based on the number of times they are generated. We use this approach to approximate the final probability distribution of the API. Our results in Section ~\ref{sec:exp} show how many queries might be enough for each stage to have a good approximation of the final probability distribution.

 \begin{figure*}[h]
     \centering
     \includegraphics[width = 0.9\linewidth]{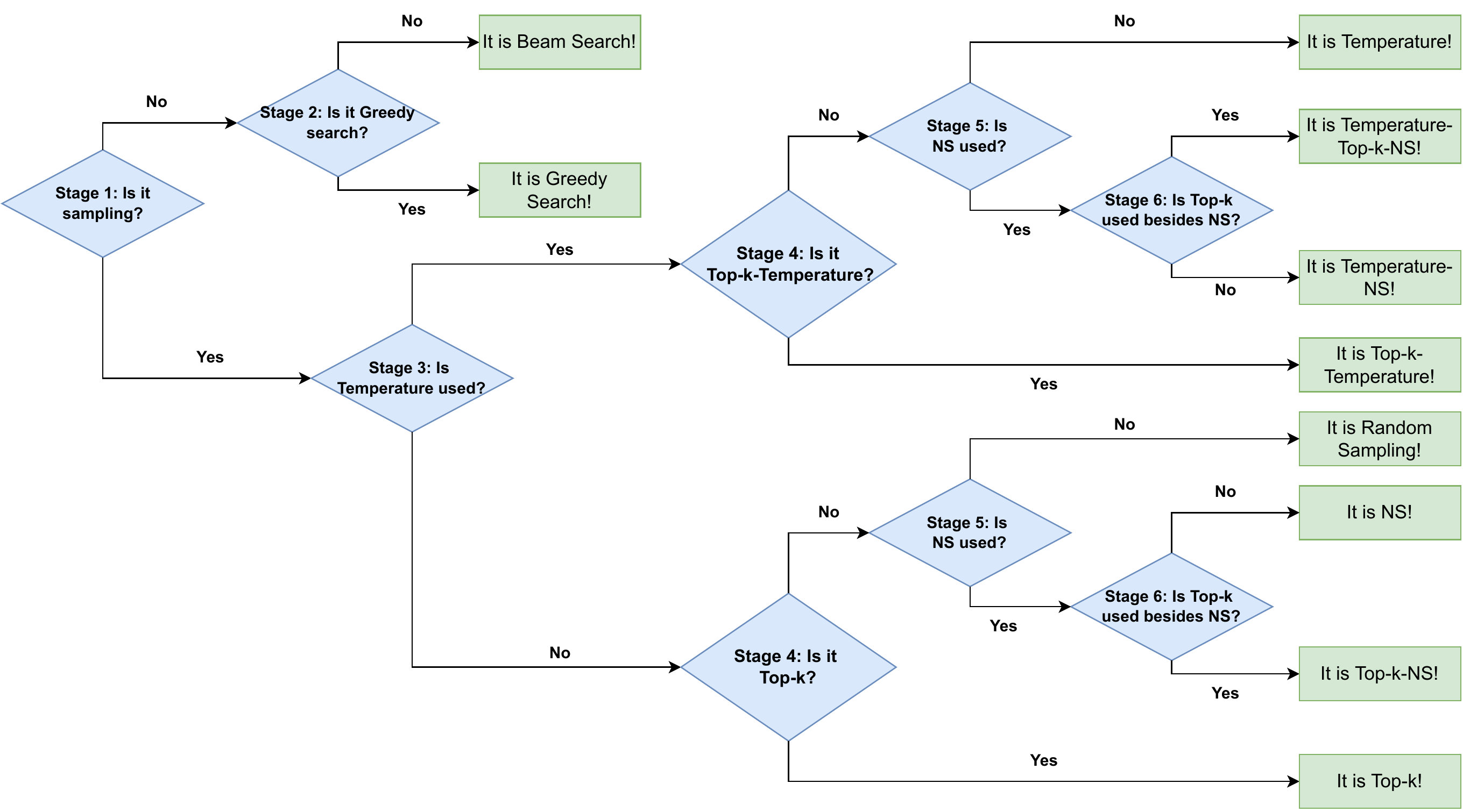}
     \vspace{-2ex}
     \caption{This flowchart presents an overview of all stages of our attack algorithm.}
     \label{fig:algorithm}
     \vspace{-2mm}
 \end{figure*}

\paragraphb{Stage 3: Does the API use temperature to decode?} As the first step towards detecting sampling-based decoding methods, we want to see if the API uses a temperature $\tau \neq 1$ to decode or not. In the next theorem, we attempt to find a formula for $\tau$.

\textbf{Theorem 1.} Assume that the victim API uses random sampling with temperature as its decoding algorithm. Suppose that $p_{1},p_{2},...,p_{|V|}$ are descending sorted inner probabilities of token among all vocabularies provided by the API's model. Also, assume that $p'_{1},p'_{2},...,p'_{|V|}$ are sorted approximated final probabilities generated by the API. Then we have $\tau = \frac{\ln (\frac{p_i}{p_j})}{\ln (\frac{p'_i}{p'_j})}$.

\vspace{-2mm}
\begin{proof} Suppose that $l_{1},l_{2},...,l_{|V|}$ are sorted logits generated by the API's model. So, we have $p_i = \frac{e^{l_i}}{\sum_{j=1}^{|V|}e^{l_j}}$ and then $p'_i = \frac{e^{\frac{l_i}{\tau}}}{\sum_{j=1}^{|V|}e^{\frac{l_j}{\tau}}}$. Set $S = \sum_{j=1}^{|V|}e^{l_j}$. So, we have $ p_i\times S = e^{\frac{l_i}{\tau}} $ and $ p_j\times S = e^{l_j} $. If we divide both side of the recent two equations, we have 

\begin{equation}
    \frac{p_i}{p_j} = \frac{e^{l_i}}{e^{l_j}} = e^{l_i-l_j} \Rightarrow l_i-l_j = \ln (\frac{p_i}{p_j})
\end{equation}

Similarly, we have 

\begin{equation}
    \frac{p'_i}{p'_j} = \frac{e^{\frac{l_i}{\tau}}}{e^{\frac{l_j}{\tau}}} = e^{\frac{l_i-l_j}{\tau}} \Rightarrow \frac{l_i-l_j}{\tau} = \ln (\frac{p'_i}{p'_j})
\end{equation}

If we substitute the equation (5), we have:

\begin{equation}
    \tau = \frac{l_i-l_j}{\frac{p'_i}{p'_j}} = \frac{\ln (\frac{p_i}{p_j})}{\ln (\frac{p'_i}{p'_j})}
\end{equation} 

\end{proof}

In the next theorem, we will demonstrate that even if the API employs temperature and Nucleus Sampling simultaneously, the formula $\tau = \frac{\ln (\frac{p_i}{p_j})}{\ln (\frac{p'_i}{p'_j})}$ remains valid. Intuitively, this is because the normalization applied after implementing Nucleus Sampling is canceled by the division of probabilities.

\textbf{Theorem 2.} Assume that the victim API uses Nucleus Sampling with the hyperparameter $p$ and temperature $\tau$ as its decoding algorithm. Suppose that $p_{1},p_{2},...,p_{|V|}$ are descending sorted inner probabilities of token among all vocabularies provided by the API's model. Also, assume that $p'_{1},p'_{2},...,p'_{|P|}$ are sorted approximated final probabilities generated by the API. Then we have $\tau = \frac{\ln (\frac{p_i}{p_j})}{\ln (\frac{p'_i}{p'_j})}$.

\begin{proof} Suppose that $l_{1},l_{2},...,l_{|V|}$ are sorted logits generated by the API's model. In this scenario, in the decoding step, the API first apply the temperature $\tau$, and then pick the minimum number of tokens whose probabilities sum exceeds $p$. Then it samples from this set of tokens. Now, assume that $p''_{1},p''_{2},...,p''_{|V|}$ are sorted probabilities after applying the temperature and before applying Nucleus Sampling. In other words, we have $p_i = \frac{e^{l_i}}{\sum_{j=1}^{|V|}e^{l_j}}$, $p''_i = \frac{e^{\frac{l_i}{\tau}}}{\sum_{j=1}^{|V|}e^{\frac{l_j}{\tau}}}$ and $p'_i = \frac{p''_i}{\sum_{j=1}^{|P|}p''_j}$ where $|P|$ is the number of tokens selected by Nucleus Sampling. Note that after applying Nucleus Sampling, the selected probabilities will get scaled to sum up to 1. So we have:

\begin{equation}
    \frac{p'_i}{p'_j} = \frac{p''_i}{p''_j} = \frac{e^{\frac{l_i}{\tau}}}{e^{\frac{l_j}{\tau}}} = e^{\frac{l_i - l_j}{\tau}} \Rightarrow \tau = \frac{l_i - l_j}{\ln (\frac{p'_i}{p'_j})} = \frac{\ln (\frac{p_i}{p_j})}{\ln (\frac{p'_i}{p'_j})}
\end{equation}

The last equation is achieved from the equation (5) in the last theorem.
\end{proof}

Similarly, we can demonstrate that if the API utilizes a combination of temperature and top-k sampling, the parameter $\tau$ can still be obtained using the same formula.

\textbf{Theorem 3.} Assume that the victim API uses top-k sampling with the hyperparameter $k$ and temperature $\tau$ as its decoding algorithm. Suppose that $p_{1},p_{2},...,p_{|V|}$ are descending sorted inner probabilities of token among all vocabularies provided by the API's model. Also, assume that $p'_{1},p'_{2},...,p'_{k}$ are sorted approximated final probabilities generated by the API. Then we have $\tau = \frac{\ln (\frac{p_i}{p_j})}{\ln (\frac{p'_i}{p'_j})}$.

\begin{proof} In this scenario, in the decoding step, the API first applies the temperature $\tau$, then selects the top $k$ tokens with the highest probabilities from the resulting distribution. It then rescales the probabilities of these $k$ selected tokens so that they sum to 1 and samples from this set of tokens. To clarify, suppose that $l_{1},l_{2},...,l_{|V|}$ are sorted logits generated by the API's model, and $p''{1},p''{2},...,p''{|V|}$ are the probabilities resulting from applying the temperature to the logits. That is, we have $p_i = \frac{e^{l_i}}{\sum{j=1}^{|V|}e^{l_j}}$, $p''i = \frac{e^{\frac{l_i}{\tau}}}{\sum{j=1}^{|V|}e^{\frac{l_j}{\tau}}}$, and $p'_i$ being the probabilities after applying top-k sampling, which is obtained by $\frac{p''_i}{\sum{j=1}^{k}p''_j}$ where $k$ is the number of tokens selected by top-k sampling. So we have:

\begin{equation}
    \frac{p'_i}{p'_j} = \frac{p''_i}{p''_j} = \frac{e^{\frac{l_i}{\tau}}}{e^{\frac{l_j}{\tau}}} = e^{\frac{l_i - l_j}{\tau}} \Rightarrow \tau = \frac{l_i - l_j}{\ln (\frac{p'_i}{p'_j})} = \frac{\ln (\frac{p_i}{p_j})}{\ln (\frac{p'_i}{p'_j})}
\end{equation}
\end{proof}

Eventually, in the most complicated scenario, when the API uses all three approaches, including temperature, top-k sampling and Nucleus Sampling, at the same time, we can use the same formula to estimate the temperature. In the following theorem, we will show why this is correct.

\textbf{Theorem 4} Assume that the API uses all three sampling methods, including temperature, top-k and Nucleus Sampling, together as its decoding algorithm. Suppose that $p_{1},p_{2},...,p_{|V|}$ are descending sorted inner probabilities of token among all vocabularies provided by the API's model. Also, assume that $p'_{1},p'_{2},...,p'_{k}$ are sorted approximated probabilities generated by the API. Then we have $\tau = \frac{\ln (\frac{p_i}{p_j})}{\ln (\frac{p'_i}{p'_j})}$.

\vspace{-2mm}
\begin{proof} In this scenario, the API employs a decoding algorithm that involves multiple steps. First, it applies a temperature $\tau$ to the logits generated by its model. Next, it selects the first $k$ tokens with the highest probabilities, based on the probabilities obtained after applying temperature. Then, it applies Nucleus Sampling, and picks the minimum number of tokens whose probabilities sum exceeds a threshold $p$. Finally, it samples from this set of tokens. Now, suppose that $l_{1},l_{2},...,l_{|V|}$ are sorted logits generated by the API's model, $p''_{1},p''_{2},...,p''_{k}$ are sorted probabilities after applying the temperature and before applying top-k sampling, and $p'''_{1},p'''_{2},...,p'''_{P}$ are sorted probabilities after applying the top-k sampling and before applying Nucleus Sampling. More precisely, we have $p_i = \frac{e^{l_i}}{\sum_{j=1}^{|V|}e^{l_j}}$, $p''_i = \frac{e^{\frac{l_i}{\tau}}}{\sum_{j=1}^{|V|}e^{\frac{l_j}{\tau}}}$, $p'''_i = \frac{p''_i}{\sum_{j=1}^{k}p''_j}$, and $p'_i = \frac{p'''_i}{\sum_{j=1}^{P}p'''_j}$ where $k$ is the hyperparameter of top-k sampling and $P$ is the number of tokens whose probability sum exceeds $p$. Note that after applying top-k and Nucleus Sampling, the selected probabilities are scaled to sum up to 1. If we set $S' = \sum_{j=1}^{|V|}e^{\frac{l_j}{\tau}}$, $S'' = \sum_{j=1}^{k}p''_j$ and $S''' = \sum_{j=1}^{k}p'''_j$, we have:

\begin{equation}
    \begin{aligned}
    \frac{p'_i}{p'_j} = \frac{\frac{p'''_i}{S'''}}{\frac{p'''_j}{S'''}} = \frac{p'''_i}{p'''_j} =  \frac{\frac{p''_i}{S''}}{\frac{p''_j}{S''}} =
     \frac{p''_i}{p''_j} =  \frac{\frac{e^{\frac{l_i}{\tau}}}{S'}}{\frac{e^{\frac{l_j}{\tau}}}{S'}} =
      \frac{e^{\frac{l_i}{\tau}}}{e^{\frac{l_j}{\tau}}} = e^{\frac{l_i - l_j}{\tau}}
    \end{aligned}
\end{equation}

Thus, we have:

\begin{equation}
    \begin{aligned}
    \tau = \frac{l_i - l_j}{\ln (\frac{p'_i}{p'_j})} = \frac{\ln (\frac{p_i}{p_j})}{\ln (\frac{p'_i}{p'_j})}
    \end{aligned}
\end{equation}

\end{proof}

As we showed in theorems, in all cases, we can use the same formula to estimate temperature. So, in the third step, we can see if the API uses temperature or not, and if so, we can find the hyperparameter too (if $\tau \approx 1$ then we can conclude the temperature is not used in decoding algorrithm).

\paragraphb{Stage 4: Is the decoding algorithm one of top-k or temperature and top-k combined?} In this step, we propose a simple approach to determine whether the API uses top-k as the last component of its decoding algorithm. This approach does not determine whether top-k is used before Nucleus Sampling, if both are employed by the API. Our approach leverages the property of top-k sampling that the API always selects the first k tokens with the highest probability. To clarify, suppose we regenerate the next token for an arbitrary sequence multiple times. If we repeat this process from scratch, we consistently obtain the same set of unique tokens. The number of unique generated tokens is the hyperparameter k for this decoding approach. Consequently, by utilizing the results obtained from previous and current stages, we can determine whether the decoding algorithm is top-k or temperature and top-k.

To be more specific, we begin with an arbitrary prompt and regenerate the following token N times. If we repeat this process with other arbitrary prompts and obtain the same number of unique tokens, it suggests with a high probability that the API employs top-k sampling as the last step of its decoding algorithm, either top-k sampling or temperature and top-k sampling. The probability that the API does not use top-k sampling and we obtain the same number of unique tokens for some arbitrary sequences is very low. We do a theoretical evaluation in Section ~\ref{sec:exp} to show how probable this happens.

\paragraphb{Stage 5: Does the API use Nucleus Sampling as part of its decoding algorithm?} In this section, we will show if the API uses Nucleus Sampling as part of its decoding algorithm or not and what its hyperparameter is. In the following theorems, we attempt to find a formula to estimate the Nucleus Sampling hyperparameter $p$.

\textbf{Theorem 5.} Assuming that the victim API employs Nucleus Sampling with the hyperparameter $p$ as its decoding algorithm, and let $p_{1},p_{2},...,p_{|V|}$ be the descending sorted inner probabilities of tokens among all vocabularies provided by the API's model. Also, assume that $p'{1},p'{2},...,p'_{|P|}$ are the sorted approximated probabilities generated by the API. Then, the ratio $\frac{p_i}{p'_i}$ serves as an estimation for p.

\vspace{-1.5ex}
\begin{proof}
 Suppose that $p_{1},p_{2},...,p_{|V|}$ are sorted inner probabilities generated by the API's model at a specific time step. Also, assume that $p'_{1},p'_{2},...,p'_{|P|}$ are sorted final probabilities generated by the API after applying Nucleus Sampling as decoding algorithm. Hence, we have:

\begin{equation}
    p'_i = \frac{p_i}{\sum_{j=1}^{|P|} p_j} \Rightarrow \sum_{j=1}^{|P|} p_j = \frac{p_i}{p'_i}
\end{equation}

We propose that $\sum_{j=1}^{|P|} p_j$ serves as an acceptable approximation for p. As per the definition of Nucleus Sampling, we select the minimum number of tokens such that the sum of their probabilities exceeds p. Therefore, $\sum_{j=1}^{|P|} p_j$ represents the sum of the probabilities of these tokens. While it may be slightly different from the actual value of p for some cases, it is still an acceptable estimation. Even if we utilize Nucleus Sampling with our estimated value, the final probability distribution will remain unchanged. This claim is further confirmed by our results in Section ~\ref{sec:exp}.

\end{proof}

The above formula provides an estimation of $p$ when the probabilities before and after applying Nucleus Sampling are known. Additionally, we can use the ratio $\frac{p_i}{p'_i}$ at stage 5 to determine whether Nucleus Sampling is employed or not. To clarify, if $\frac{p_i}{p'_i} \neq 1$, it indicates that neither top-k nor Nucleus Sampling is utilized in the decoding algorithm. Once the type of sampling is determined, this formula can be used to estimate p.

For instance, after determining that the API applies temperature and Nucleus Sampling in combination, we first use equation (8) to estimate the temperature, then apply this temperature and provide a probability distribution of tokens $p_{1},p_{2},...,p_{|V|}$. After that, we can use equation (12) to estimate the hyperparameter $p$.

\paragraphb{Second approach for estimating p in Nucleus Sampling.} In this section, we propose an alternative approach for estimating the value of p in Nucleus Sampling that is more straightforward. While the first approach yields accurate estimates, it is highly dependent on the probabilities of the victim model, and small changes in these probabilities can lead to inaccurate estimates. To reduce the reliance on these probabilities, we propose the following approach.

Suppose that $p_1, p_2, ..., p_{|V|}$ are the inner probabilities generated by the API's model, and $p'_1, p'_2, ..., p'_{|P|}$ are the final probabilities generated by the API. In this case, we can estimate $p$ as $\sum_{i=1}^{|P|} p_i$. While this approach may be less accurate than the first method, it may be useful in situations where the exact inner probabilities are not available.

\paragraphb{Stage 6: Is top-k used before Nucleus Sampling?} In the final cases, we aim to investigate the potential solutions when the API employs top-k sampling before Nucleus Sampling. In the previous stage, we determined whether Nucleus Sampling is employed or not. Suppose that top-k is utilized before Nucleus Sampling in the decoding algorithm. In this scenario, if we apply formula (12) for consecutive time steps, the value obtained from the formula will vary. In other words, top-k truncates the probability distribution prior to applying Nucleus Sampling. As different probability distributions exist across different time steps, the resulting value from the formula will vary.
If temperature is also utilized (as determined in stage 3), we can apply it first and then proceed to determine if top-k is employed before Nucleus Sampling. It is important to note that this case emerges as the most complicated one in our analysis and, arguably, not very practical. A summary of our algorithm is presented in Figure \ref{fig:algorithm}.

\paragraphe{Estimating  $k$ and $p$ when they are used together:}
In stage 6 of our investigation, we determine whether the top-k sampling method is applied prior to Nucleus Sampling. If this is indeed the case, we propose a systematic approach for estimating the values of $p$ and $k$ when top-k sampling and Nucleus Sampling are employed concurrently as components of a decoding algorithm. Suppose that 
$p_{1},p_{2},...,p_{|V|}$ and $q_{1},q_{2},...,q_{|V|}$ are descending sorted inner probabilities of a token among all vocabularies provided by the API's model at time steps $t_1$ and $t_2$ respectively. Also, assume that $p'_{1},p'_{2},...,p'_{|P|}$ and $q'_{1},q'_{2},...,q'_{|P|}$are sorted approximated probabilities generated by the API at these time steps. Set $S_{k}^{(1)}=\sum_{j=1}^{k} p_{j}$, $S_{k}^{(2)}=\sum_{j=1}^{k} q_{i}$, $S_{p}^{(1)}=\sum_{j=1}^{|P|} p''_{j}$, and $S_{p}^{(2)}=\sum_{j=1}^{|P|} q''_{j}$. where $p''_{1},p''_{2},...,p''_{k}$ and $q''_{1},q''_{2},...,q''_{k}$ are sorted probabilities after applying the top-k and before applying Nucleus Sampling. It is not so hard to show that $p'_{i} = \frac{p_i}{S_{k}^{(1)}S_{p}^{(1)}}$ and $q'_{i} = \frac{q_i}{S_{k}^{(2)}S_{p}^{(2)}}$.

Since $p$ is fixed, we can assume that $S_p$ is also fixed. Thus, it can be canceled from the equations. Hence, we have $p'_i = \frac{p_i}{S_{k}^{(1)}}$ and $q'_i = \frac{q_i}{S_{k}^{(2)}}$. By dividing the both sides of the recent equations, we have:

\begin{equation}
    \frac{S_{k}^{(1)}}{S_{k}^{(2)}} = \frac{p'_i}{q'_i} \times \frac{q_i}{p_i}
\end{equation}

Since there are two unknown variables here ($S_{k}^{(1)}$ and $S_{k}^{(2)}$), we set the value for one of them to estimate the other one. To accomplish this, we begin by considering different values of k, and then calculate $S_{k}^{(1)}$ at time step $t_1$. Using the formula (13), we can compute $S_{k}^{(2)}$ for time step $t_2$. Then, we can examine the corresponding value of k, which leads to the sum $S_{k}^{(2)}$ in the probability distribution of time step $t_2$. We repeat this procedure until we find a k such that the corresponding hyperparameter k' resulting from the second time step equals k. Next, we use the estimated k to provide the new probability distribution resulting from top-k, and then apply formula (12) to estimate p. Our experiments will demonstrate the accuracy of the estimation resulting from this approach.

\section{Evaluations and Experiments }\label{sec:exp}

In this study, we conduct separate experiments on GPT-2, GPT-3, and GPT-Neo models due to the extensive utilization of these models across numerous APIs in various applications. For the experiments on GPT-2 and GPT-Neo, we use the Huggingface\footnote{https://huggingface.co} library, which allows us to implement various combinations of decoding algorithms. On the other hand, for the experiments on GPT-3, we use the OpenAI API, which only provides access to temperature and Nucleus Sampling. While our proposed attacks are agnostic to the size of the underlying LM, due to the large number of experiments and computational constraints we primarily conduct our experiments on the smaller versions of both models. Nevertheless, in Appendix ~\ref{sec:size_effect}, we conduct some representative experiments on larger GPT-2 (medium, large) and GPT-3 models (babbage and curie) to show generalization of our proposed attacks across model sizes. 
Furthermore, we perform some experiments utilizing the GPT-Neo model. Also, it should be noted that our experiments involve utilizing a custom decoding algorithm in conjunction with GPT-2/3/Neo, rather than directly attacking an actual deployment of these language models. The code used to conduct the experiments presented in this paper is available.\footnote{\url{https://github.com/SPIN-UMass/Stealing-the-Decoding-Algorithms-of-Language-Models}}

\begin{table*}[t]
    \begin{center}
    \caption{Results of temperature estimation with different decoding combinations.  The accuracy of the estimation is measured using the K-S test p-value and KL divergence score, which indicate the similarity between the provided probability distributions using the estimated hyperparameters and the actual hyperparameters.}
    \vspace{-2ex}
    \label{tab:temperature2}
    \resizebox{0.93\textwidth}{!}{
    \begin{tabular}{ cccccc } 
     \toprule
    Decoding Strategy & Real Temperature & Estimated Temperature & p-value & KL Divergence\\
    \midrule
    Top-k ($k=30$) \& Temperature & 0.85 & 0.8568 $\pm$ 0.016 & 1.0 & 0.002 $\pm$ 0.016\\
    Top-k ($k=40$) \& Temperature & 0.75 & 0.7569 $\pm$ 0.016 & 1.0 & 0.007 $\pm$ 0.011\\
    Top-k ($k=60$) \& Temperature & 0.65 & 0.6586 $\pm$ 0.014 & 1.0 & 0.002 $\pm$ 0.009\\
    \midrule
    NS ($p=0.9$) \& Temperature & 0.85 & 0.8575 $\pm$ 0.018 & 1.0 & 0.004 $\pm$ 0.01\\
    NS ($p=0.8$) \& Temperature & 0.8  & 0.8080 $\pm$ 0.014 & 1.0 & 0.008 $\pm$ 0.01\\
    NS ($p=0.85$) \& Temperature & 0.75 & 0.7579 $\pm$ 0.017 & 1.0 & 0.007 $\pm$ 0.016\\
    \midrule
    Top-k ($k=40$) \& NS ($p=0.8$) \& Temperature  & 0.9 & 0.9096 $\pm$ 0.017 & 0.996 & 0.001 $\pm$ 0.0163\\
    Top-k ($k=50$) \& NS ($p=0.8$) \& Temperature  & 0.85 & 0.8603 $\pm$ 0.017 & 1.0 & 0.003 $\pm$ 0.012\\
    Top-k ($k=60$) \& NS ($p=0.8$) \& Temperature  & 0.80 & 0.8022 $\pm$ 0.014 & 1.0 & 0.002 $\pm$ 0.01\\
    \bottomrule
    \end{tabular}
    }
    \end{center}
    \vspace{-2mm}
\end{table*}

\subsection{Evaluation Metrics}
To determine the similarity in functionality between two LMs or text generation-based APIs, we must compare the probability distributions generated by them at each time step. This is because similar decoding algorithms and corresponding hyperparameters will result in similar truncated probability distributions. To evaluate our estimation, we compare the probability distributions of the victim API and the API that uses our estimation as the type and corresponding hyperparameters. In this paper, we use two metrics to achieve this goal: the Kolmogorov-Smirnov test~\cite{massey1951kolmogorov} and the Kullback-Leibler divergence (KL divergence)~\cite{kullback1951information}. 

\textbf{Kolmogorov-Smirnov Test:} One such metric is the Kolmogorov-Smirnov test (K-S test). The K-S test is a non-parametric statistical test that is used to compare two discrete probability distributions. The test evaluates the similarity between the two distributions by comparing the cumulative distribution functions (CDFs) of the observed and hypothesized distributions.
The K-S test calculates the maximum distance (referred to as the K-S statistic) between the two CDFs and compares it to a critical value determined by the sample size. If the calculated K-S statistic is larger than the critical value, the null hypothesis that the two distributions are the same is rejected. Results of the K-S test are typically reported as a p-value, which represents the probability that the observed differences between the two distributions are due to chance. A small p-value (typically less than 0.05) indicates that the observed differences are statistically significant and that the two distributions are likely different from each other.

\textbf{KL Divergence:} Another metric for comparing probability distributions is the KL divergence. KL divergence is a measure of the difference between two probability distributions. It is defined as the sum of the product of the probability of each event in one distribution with the natural logarithm of the ratio of the probability of that event in the other distribution. KL divergence is a non-negative value, and it is zero only if the two distributions are identical.

Given that the KL-divergence metric yields a single value, it is necessary to determine what is a good KL-divergence score in our specific context. To address this concern, we compare numerous pairs of probability distributions generated by using the same decoding type and hyperparameters. The average score for these pairs is $0.0017 \pm 0.0007$. This value can serve as a benchmark for evaluation. However, it should be noted that KL-divergence scores typically exceed $0.1$ when different hyperparameters are used.

It is worth noting that MAUVE~\cite{pillutla2021mauve} is a recently proposed automatic metric for evaluating language generators. In other words, it measures the similarity of token distribution between two generated texts. However, like KL-divergence, it also produces a single scalar value. Additionally, evaluating each pair of decoding algorithms using MAUVE requires generating a large number of lengthy sequences using each decoding algorithm, which is not feasible for the number of experiments conducted in this study.

\subsection{Evaluation of Each Stage on GPT-2}
Since each stage of our attack uses a different algorithm, here we evaluate the performance of each stage of our attack, described in Section ~\ref{subsec:attack}, one by one. 

\paragraphb{Stage 1:} As theoretical evaluation for this stage, assume the API uses a sampling-based method. The API rarely generates the same sequence if we query the API many times. More precisely, suppose we use the API to generate a sequence of length $50$ each time. Also, assume that $p_1,p_2,...,p_{50}$ are the probabilities that each tokens can be generated. Thus, if we send the same prompt as query $N$ times, the probability that API generates the same sequence is $p_{1}^{N} \times p_{1}^{N} \times ... \times p_{1}^{N}$. Even if we set $p_i = 0.99$ and $N=20$, this probability will be around $4.31e-5$ which is very low. So, it always predicts correctly.

\paragraphb{Stage 2:}
In the second stage of our evaluation, we consider 1000 different APIs. Each API may randomly use either greedy search or beam search as its decoding algorithm. Additionally, if the API uses beam search, the beam size is chosen randomly from a range of common values, specifically between 2 and 10. Our results show that we can detect the type of decoding method used with $100\%$ accuracy. Furthermore, our experiments demonstrate that we can detect the beam size with a high degree of accuracy, using only 40 arbitrary prompts. While it is not straightforward to provide a theoretical justification for the number of queries required to guarantee the correct detection of the beam size, we provide examples in the Appendix ~\ref{sec:beam_size} that support our results.  

\begin{figure}[t]

\centering
\includegraphics[width=8.0cm,height=\textheight,keepaspectratio]{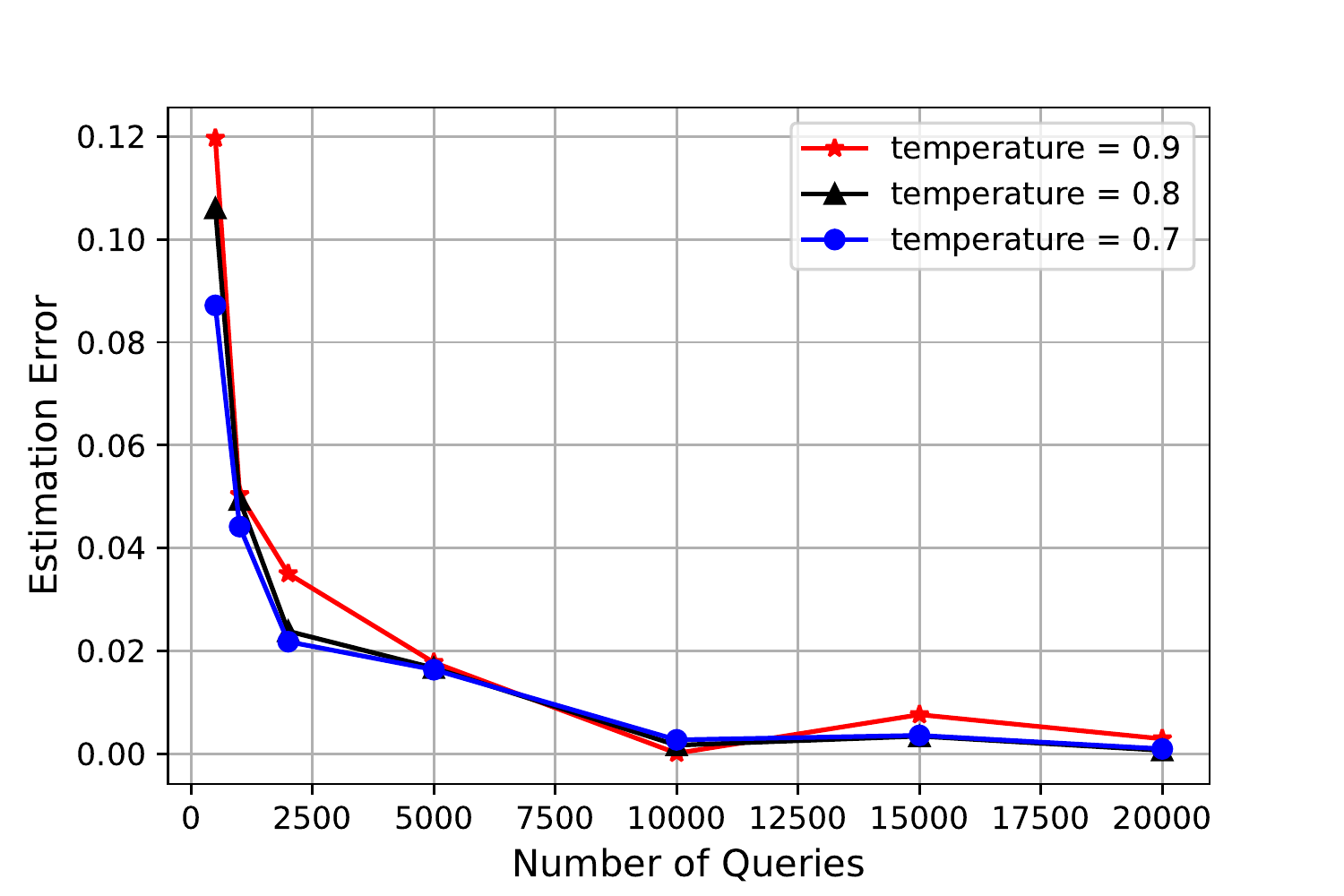}
\vspace{-2ex}
\caption{Estimation error of the temperature $\tau$ for different numbers of queries. The graph illustrates that using $10000$ queries may be sufficient to achieve an accurate estimation.}
\label{fig:temp}
\vspace{-2mm}
\end{figure}

\paragraphb{Stage 3:}
In the third stage, we estimate the temperature for scenarios where it is used as part of the decoding algorithm. We use the inner and final probabilities generated by the API and apply the formula (7) to estimate the temperature. To approximate the final probabilities generated by the API, we send an arbitrary prompt to the API and generate the next token $N$ times. By increasing $N$, we can achieve a more accurate estimation. In our experiment, we use the arbitrary sequence "Students opened their" as a prompt. Figure ~\ref{fig:temp} illustrates the number of queries required to achieve an accurate estimation of temperature. Additionally, we consider different cases where temperature is used as part of the decoding method and show that the same formula can be used to estimate the temperature in all cases. To demonstrate this, we use various arbitrary values of hyperparameters $k$ and $p$ for top-k and Nucleus Sampling respectively (Table~\ref{tab:temperature2}).

As shown in Figure ~\ref{fig:temp}, with only $1000$ queries, we cannot get an acceptable estimation. However, increasing the number of queries makes the estimation more accurate. Even with $10000$ queries, we can estimate the temperature.

\paragraphb{Stage 4:}
In the fourth stage, we aim to determine whether the decoding algorithm used by the API employs top-k sampling as its last component, and if so, estimate the value of the hyperparameter $k$. To do this, we select four arbitrary prompts and repeatedly generate the next token $N$ times, where $N$ is chosen based on the expected value of $k$. If the number of unique tokens generated across all sequences is the same, it is likely that the API's decoding algorithm uses top-k sampling, and the number of unique tokens will serve as our estimate for the value of $k$. We apply this algorithm to eight different decoding strategies and for a range of values for $k$ from 10 to 100. The results in Table ~\ref{tab:topk} indicate the number of queries used in our experiments to estimate $k$. It should be noted that these values do not necessarily represent the minimum number of queries required to obtain an accurate estimate of $k$.

Additionally, it is important to consider the number of queries needed to ensure that all top-k tokens have been generated at least once, including the least probable token among them. This question can be viewed as a variant of the well-known problem of determining the expected number of trials until success, where success is defined as the generation of the least probable top-k token. If the probability of success is $p$, the expected number of trials until success is $\frac{1}{p}$. By using the inner probability distribution provided by the API's model, we can gain insight into the lower bound for the number of queries needed. In practice, the attacker may choose to send more queries than this lower bound to ensure an accurate estimation.

As previously discussed in Section ~\ref{sec:Stealing Attack}, this approach is ineffective if there is at least one token among the top-k tokens that the API will not generate. The probability of this occurrence is $(1-p)^N$, where $p$ is the probability of that token being generated by the API's model. As an example, if $p=0.0001$ for a token, and we send queries $N=50000$ times, we will have $(1-p)^N = 6.7\times 10^{-3}$. To decrease the likelihood of this approach failing, we must either increase $N$ or $p$. Increasing $N$ may be cost-prohibitive, so to increase $p$, we can select prompts that result in a more flattened probability distribution over the next token. We use Kurtosis as a metric to evaluate the level of flattening in the probability distribution.

It is also important to note that when using temperature in addition to top-k sampling, the probability of less likely tokens being generated decreases. This means that more queries are needed to ensure they will be generated in order to make an accurate prediction for the hyperparameter $k$.

\begin{figure}[t]
\centering
\includegraphics[width=8.0cm,height=\textheight,keepaspectratio]{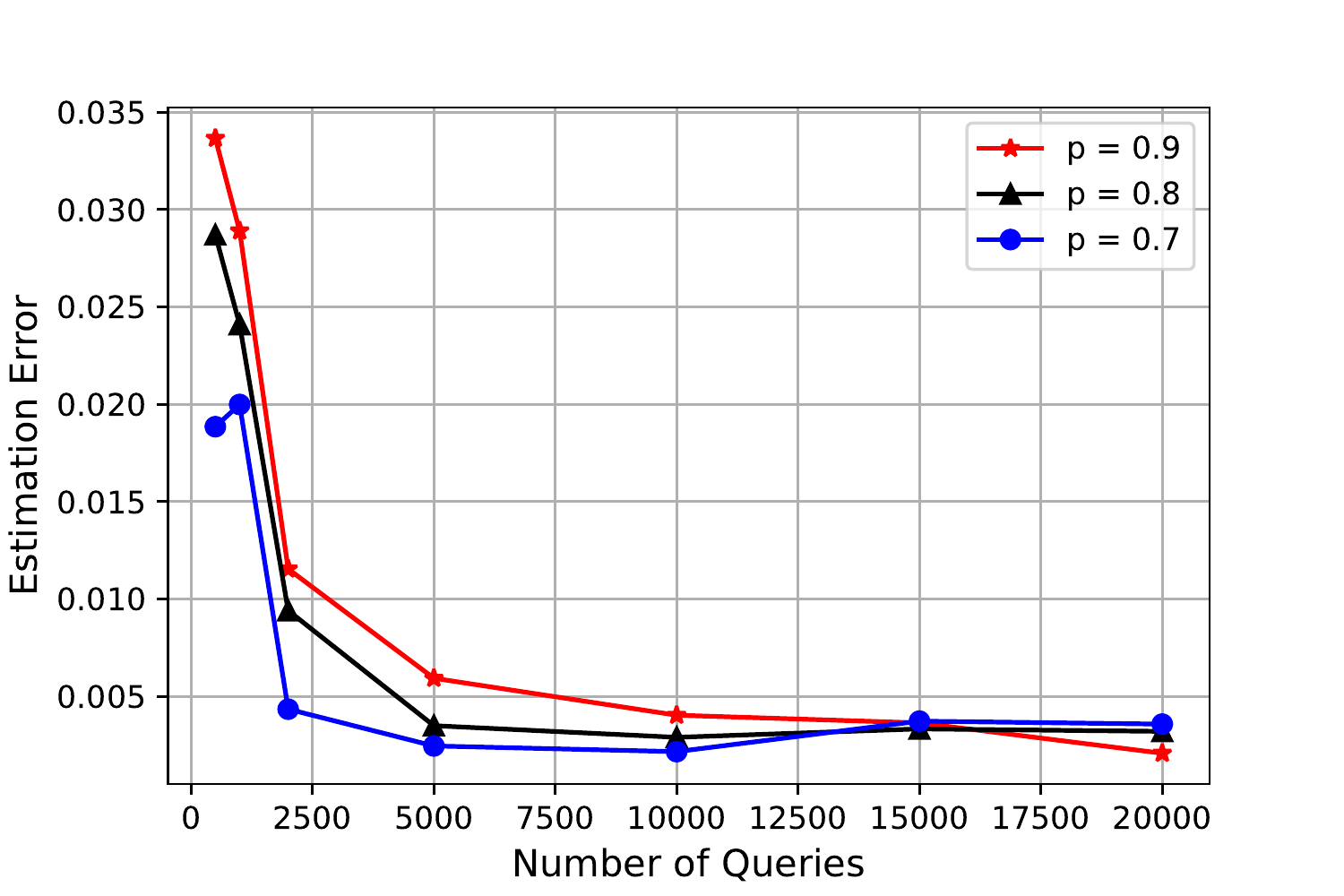}
\vspace{-2ex}
\caption{Estimation error of the hyperparameter $p$ for different numbers of queries. The graph shows that even with as few as $5000$ queries, an accurate estimation of the hyperparameter $p$ can be achieved.}
\label{fig:ns_queries}
\vspace{-5mm}
\end{figure}

\paragraphb{Stage 5:}
In order to evaluate this stage, we will only consider scenarios that involve Nucleus Sampling and Nucleus Sampling with temperature. As previously discussed in Section ~\ref{sec:Stealing Attack}, it is not possible to use Equation (12) to estimate the hyperparameter $p$ when the API also utilizes top-k sampling. This will be addressed in the next stage. However, we can use Equation (12) to confirm that Nucleus Sampling is being used (if $p$ computed by Equation (12) does not equal 1). We calculate Equation (12) for multiple tokens to obtain a more reliable estimation, and then take the average. Figure ~\ref{fig:ns_queries} illustrates the estimation and the required number of queries, and Table ~\ref{tab:NS} presents the results for various decoding strategies.

\paragraphb{Stage 6:}
In the last stage, we aim to figure out if top-k sampling is used before Nucleus Sampling and then estimate $p$ and $k$ (if top-k is used). To do so, as described in Section ~\ref{sec:Stealing Attack}, we need the final probability distributions in two different time steps. Hence, we use two different prompts: "My school is close to the" and "Students opened their". Then we apply the formula (13) to estimate $k$. After finding $k$, then we can apply the formula (12) to estimate $p$. Results shown in Table ~\ref{tab:Topk_NS}.

\begin{table}[t]
    \begin{center}
    \caption{This table displays the query count required for a precise estimation of the hyperparameter $k$, which may be slightly higher  than the minimal necessary number.}
    \vspace{-2ex}
    \label{tab:topk}

    \resizebox{0.48\textwidth}{!}{
    \begin{tabular}{ ccc } 
     \toprule
    Decoding Strategy & Real k & Number of Queries \\
    \midrule
    Top-k & 30 & 700  \\
    Top-k & 40 & 1,000  \\
    Top-k & 60 & 1,500  \\
    Top-k & 80 & 4,000  \\
    Top-k & 100 & 8,000  \\
    \midrule
    Top-k \& Temperature ($\tau=0.8$) & 30 & 1,500  \\
    Top-k \& Temperature ($\tau=0.8$) & 40 & 2,000  \\
    Top-k \& Temperature ($\tau=0.8$) & 60 & 5,000  \\
    Top-k \& Temperature ($\tau=0.8$) & 80 & 15,000  \\
    Top-k \& Temperature ($\tau=0.8$) & 100 & 20,000  \\
    \bottomrule
    \end{tabular}
    }
    \end{center}
    \vspace{-3mm}
\end{table}

\begin{table}[t]
    \begin{center}
    \caption{Results of hyperparameter $p$ estimation employing strategies of Nucleus Sampling (NS) alone or combined with temperature.}
    \vspace{-2ex}
    \label{tab:NS}

    \resizebox{0.48\textwidth}{!}{
    \begin{tabular}{ cccc } 
     \toprule
    Decoding Strategy & Real p & Estimated p & p-value\\
    \midrule
    NS & 0.75 & 0.75 $\pm$ 0.005 & 1.0\\
    NS & 0.90 & 0.899 $\pm$ 0.005 & 1.0\\
    NS & 0.85 & 0.847 $\pm$ 0.005 & 0.99 $\pm$ 0.011\\
    Temperature ($\tau=0.85$) \& NS & 0.80 & 0.801 & 1.0\\
    Temperature ($\tau=0.75$) \& NS & 0.70 & 0.699 & 0.99 $\pm$ 0.009\\
    Temperature ($\tau=0.65$) \& NS & 0.70 & 0.696 & 1.0\\
    \bottomrule
    \end{tabular}
    }
    \end{center}
    \vspace{-3mm}
\end{table}

\paragraphb{End-to-end Analysis of the Attack:}
We consider all these steps as a single framework in the second part of our experiments. In other words, we consider 100 different APIs randomly picks one of discussed decoding strategies scenario with random values as their hyperparameters. Then, we apply our algorithm and see which combination of decoding algorithms the API uses. The whole API's decoding algorithms have been predicted correctly by our attack.

 \subsection{Experiments on GPT-3}

 Unlike HuggingFace, which offers a variety of options for utilizing the GPT-2 model, OpenAI's functionality is more limited. Specifically, the text completion function provided by OpenAI includes options for controlling the diversity of the output, such as temperature and top-p for Nucleus Sampling. However, it does not offer features such as top-k or beam search. Given this constraint, we decided to conduct experiments with GPT-3 in a separate section. We also discovered that in the generation function, employing both temperature and top-p results in only the temperature parameter being applied. Therefore, it is generally recommended to set either top-p or temperature to 1 when using one of these options.

 Our experiments involving GPT-3 were executed in a manner analogous to those conducted with GPT-2. OpenAI facilitates access to the GPT-3 API via Python code, employing a public key for querying purposes. This approach allows us to derive the final probability distribution through multiple API queries. The GPT-3 API supplies the inner probability, which we then utilize in conjunction with the formula proposed in our paper to estimate the hyperparameters effectively.

 In order to determine which decoding algorithm has been employed by the API, we must first identify whether temperature or Nucleus Sampling is being used. To do this, we evaluate the temperature formula and check if it is equal to 1, which indicates that temperature is not being utilized. In this case, either top-p is being used or neither temperature nor top-p are being utilized. To determine whether top-p is being used, we apply the formula for top-p and check if Nucleus Sampling is being applied. Through this process, we are able to detect the type of decoding algorithm as well as its corresponding value.

 In these experiments, we set the initial prompt as "My school is close to" and generated the next two tokens 10000 times. We then applied the corresponding formulas to both the original probability distribution provided by the OpenAI API and the resulting distribution obtained by querying the victim API. In order to increase the reliability of our results, we repeated these experiments 8 times and present the results in the Table ~\ref{tab:gpt3}.

 \subsection{Experiments on GPT-Neo} \label{sec:gpt-neo}

As previously mentioned, GPT-Neo is a language model similar to GPT-3 that is pre-trained on the Pile dataset. Developed by EleutherAI, the Pile is a large-scale, diverse, and high-quality dataset designed for training language models. In this section, we conduct several experiments using GPT-Neo 1.3B, which has 1.3 billion parameters. The experimental settings remain the same, with 10,000 queries sent each time, and the process is repeated four times to obtain more consistent estimations. The results are presented in Table ~\ref{tab:gpt-neo1} and ~\ref{tab:gpt-neo2}.

These results effectively illustrate the close alignment of hyperparameter estimations across various scenarios for the GPT-Neo model when employed by the API. By analyzing these results, it becomes apparent that our estimation methods for GPT-Neo yield results that are on par with those of GPT-2 and GPT-3, thus confirming the validity and robustness of our approach in accurately estimating the hyperparameters of the decoding algorithm for GPT-Neo.

\begin{table}[t]
    \begin{center}
    \caption{Hyperparameter estimation results for a combined top-k and Nucleus Sampling (NS) decoding strategy, focusing on the hyperparameters $k$ and $p$.}
    \vspace{-2ex}
    \label{tab:Topk_NS}

    \resizebox{0.48\textwidth}{!}{
    \begin{tabular}{ cccc } 
     \toprule
    Decoding Strategy & Estimated k & Estimated p & p-value \\
    \midrule
    NS ($p=0.80$) \& Top-k ($k=30$) & 28 & 0.824 & 0.96 $\pm$  0.043\\
    NS ($p=0.90$) \& Top-k ($k=30$) & 29 & 0.916 & 0.99 $\pm$ 0.005 \\
    NS ($p=0.80$) \& Top-k ($k=40$) & 35 & 0.833 & 0.97 $\pm$ 0.023 \\
    NS ($p=0.90$) \& Top-k ($k=40$) & 38 & 0.909 & 0.99 $\pm$ 0.004 \\
    NS ($p=0.80$) \& Top-k ($k=50$) & 49 & 0.806 & 0.99 $\pm$ 0.004 \\
    NS ($p=0.90$) \& Top-k ($k=50$) & 48 & 0.909 & 0.99 $\pm$ 0.01 \\
    \bottomrule
    \end{tabular}
    }
    \end{center}
    \vspace{-3mm}
\end{table}

\begin{table}[t]
    \begin{center}
    \caption{Hyperparameter estimation in a GPT-3-based API utilizing either Nucleus Sampling (NS) or temperature decoding.}
    \vspace{-2ex}
    \label{tab:gpt3}

    \resizebox{0.48\textwidth}{!}{
    \begin{tabular}{ cccc } 
     \toprule
    Decoding Strategy & Real Value & Estimated Value & p-value\\
    \midrule
    Temperature & 0.6 & 0.596 $\pm$  0.005 & 1.0\\
    Temperature & 0.7 & 0.701 $\pm$  0.004 & 1.0 \\
    Temperature & 0.8 & 0.795 $\pm$  0.006 & 1.0 \\
    \midrule
    NS & 0.6 & 0.601 $\pm$  0.001 & 1.0 \\
    NS & 0.8 & 0.801 $\pm$  0.007 & 1.0 \\
    NS & 0.9 & 0.903 $\pm$  0.006 & 1.0 \\
    \bottomrule
    \end{tabular}
    }
    \end{center}
    \vspace{-3mm}
\end{table}

\subsection{Attack Efficacy in an API with Prompt Engineering} \label{sec:other_discussion}

As previously discussed, the only knowledge the attacker requires is the probabilities of the top two tokens generated by the API's LM. There are multiple ways in which an attacker can acquire this information, which are described in detail in Section ~\ref{sec:Capabilities}. One option is to use a model stealing approach to obtain the internal probability distribution. However, since there have been no successful model-stealing attacks proposed for text generation tasks, the attacker may be motivated to employ our attack without direct access to the probabilities. In this section, we will demonstrate how the attacker can do this when the API relies on prompt engineering rather than fine-tuning.

GPT-2/3/Neo models can be applied in various applications through direct usage, prompt engineering, or fine-tuning. Fine-tuning these models for downstream tasks requires computational resources and a private dataset. These limitations have motivated the use of prompt engineering to generate desired text in many text-based tools and applications. There are various platforms that allow users to generate high-quality written content such as blog posts, product descriptions, and other personalized content, such as OpenAI playground, ChatGPT\footnote{https://chat.openai.com}, Articoolo\footnote{http://articoolo.com}, and Perplexity AI\footnote{https://www.perplexity.ai}.

In these cases, the attacker can use long prompts to generate probability distributions that are similar to those of the victim model. Specifically, by querying the API with a long prompt, the attacker can use the same prompt and the base model as a reference to approximate the internal probability distribution. Our experiments have shown that adding a prompt to the beginning of a long text does not significantly change the probability distribution over the next token. It is important to note that the long text must be in the same context. For example, if the attacker is attempting to attack a story generation API, using a long drama story and asking the API to complete it as a drama story will result in a probability distribution that is not vastly different. More detailed results are provided in the Appendix ~\ref{sec:prompt_eng}.

\begin{table*}[t]
    \begin{center}
    \caption{Results of hyperparameter estimation for GPT-Neo. This table presents the results of our estimation for the hyperparameters $\tau$ and $p$ when the API uses GPT-Neo 1.3B models.}
    \label{tab:gpt-neo1}
    \begin{tabular}{ c|cccc } 
     \toprule
    Decoding Strategy & Real Value & Estimated Value & p-value & KL Divergence \\
    \midrule
    \multirow{4}{*}{Temperature} & 0.7 & 0.7132 $\pm$ 0.021 & 1.0 & 0.003 $\pm$ 0.002 \\
     \cmidrule{2-5}
     & 0.8 & 0.8195 $\pm$ 0.035 & 0.99 $\pm$ 0.008 & 0.002 $\pm$ 0.011 \\
     \cmidrule{2-5}
     & 0.9 & 0.9154 $\pm$ 0.066 & 1.0 & 0.006 $\pm$ 0.003 \\
    \midrule
    \multirow{4}{*}{Nucleus Sampling} & 0.7 & 0.706 $\pm$ 0.011 & 1.0 & 0.006 $\pm$ 0.01 \\
     \cmidrule{2-5}
     & 0.8 & 0.8056 $\pm$ 0.016 & 0.97 $\pm$ 0.006 & 0.006 $\pm$ 0.012 \\
     \cmidrule{2-5}
     & 0.9 & 0.9104 $\pm$ 0.017 & 1.0  & 0.009 $\pm$ 0.002 \\
    \bottomrule
    \end{tabular}
    \end{center}
    
\end{table*}
\begin{table*}[t]
    \begin{center}
    \caption{Temperature estimation for the GPT-Neo 1.3B model with various decoding combinations. The K-S test p-value and KL divergence score assess the estimation accuracy by comparing the derived and actual hyperparameter distributions.}
    \vspace{-2ex}
    \label{tab:gpt-neo2}
    \resizebox{0.93\textwidth}{!}{
    \begin{tabular}{ cccccc } 
     \toprule
    Decoding Strategy & Real Temperature & Estimated Temperature & p-value & KL Divergence\\
    \midrule
    Top-k ($k=30$) \& Temperature & 0.85 & 0.8659 $\pm$ 0.033 & 1.0 & 0.001 $\pm$ 0.001\\
    Top-k ($k=40$) \& Temperature & 0.75 & 0.7563 $\pm$ 0.021 & 1.0 & 0.003 $\pm$ 0.001\\
    Top-k ($k=60$) \& Temperature & 0.65 & 0.6561 $\pm$ 0.015 & 1.0 & 0.001 $\pm$ 0.001\\
    \midrule
    NS ($p=0.9$) \& Temperature & 0.85 & 0.8665 $\pm$ 0.037 & 1.0 & 0.004 $\pm$ 0.01\\
    NS ($p=0.8$) \& Temperature & 0.8  & 0.8097 $\pm$ 0.025 & 0.992 & 0.008 $\pm$ 0.013\\
    NS ($p=0.85$) \& Temperature & 0.75 & 0.7567 $\pm$ 0.019 & 1.0 & 0.007 $\pm$ 0.016\\
    \bottomrule
    \end{tabular}
    }
    \end{center}
    \vspace{-2mm}
\end{table*}

\section{Analysis of the Cost of the Attack}\label{sec:Cost}

In this section, we aim to provide an estimation of the costs associated with querying the API in order to execute our hyperparameter stealing algorithm. Our primary motivation for this attack is that hiring individuals for human evaluation to determine the best decoding algorithm and corresponding hyperparameters can be costly. Therefore, stealing this information at a lower cost is desirable.

To analyze the cost of our algorithm, we consider the worst-case scenario. In the worst case, if the API utilizes both top-k and Nucleus Sampling as its decoding algorithm, as depicted in Figure ~\ref{fig:algorithm}, all stages of our algorithm must be executed. To do so, we need to send approximately $400,000$ queries to the API. Each query consists of $5$ tokens. As a result, we will have $2,000,000$ tokens in total. Therefore, in the worst-case scenario, the cost of querying the API with $2,000,000$ tokens would be $ (2,000,000/1000 \times x)$, where $x$ is the cost of querying the API with 1000 tokens.

As an example, OpenAI provides pricing information for different versions of the GPT-3 API. GPT-3 has four versions, named Ada, Babbage, Curie, and Davinci. The cost for these models are $\$0.0004$, $\$0.0005$, $\$0.002$, and $\$0.02$ per 1000 tokens, respectively. Ada is the fastest and cheapest version, while Davinci is the most powerful and expensive one. Therefore, the cost for this API would be $\$0.8$, $\$1$, $\$4$, and $\$40$ for the four versions respectively.

\section{Potential Countermeasure}

This paper demonstrates that LM-based APIs are vulnerable to hyperparameter stealing attacks, specifically, that the information in the decoding section of LMs in open-ended text generation tasks can be extracted. This highlights the need for APIs to develop defense mechanisms against such attacks. These attacks are dependent on the accuracy of the API's final probability distribution. As a defense, the API can introduce noise into this probability distribution by randomly replacing generated tokens at certain time steps with a probability of $0.1$.

A similar approach, known as watermarking, was introduced in~\cite{szyller2021dawn}. By introducing this noise, the final probability distribution is different from the original one, thus making it more difficult for an attacker to extract the hyperparameters using the formula proposed in the paper. This method is particularly effective for sensitive hyperparameters such as temperature, as small changes in temperature can lead to significant changes in the probability distribution. Table ~\ref{tab:Countermeasure} shows that this method is able to effectively defend against the attack.

It should be noted that any defense mechanism may hurt the API's performance. To minimize this impact, the API can replace the generated token with a random one among the most probable tokens. This will disrupt the attacker's ability to extract the hyperparameters while minimizing the effect on the API's performance.

However, to demonstrate that our proposed defense does not negatively impact performance, we select 150 samples from our genre-based story generation dataset and provide 150 corresponding prompts for the text completion task. We then employ both the modified text generation system and the system without our defense to complete these prompts. We utilize perplexity as a metric to illustrate that the performance of our proposed defense is not significantly affected. Table ~\ref{tab:Countermeasure} presents the perplexity of the generated text both with and without our proposed defense.

\begin{table*}[t]
    \begin{center}
    \caption{Results of hyperparameter estimation after applying proposed countermeasure. The table shows the p-values of the new estimations, which confirm that the ruined estimations lead to a significantly different probability distribution. Additionally, the table compares system perplexity before and after deployment, indicating a minor alteration.}
    \label{tab:Countermeasure}

    \resizebox{0.93\textwidth}{!}{
    \begin{tabular}{ ccccc } 
     \toprule
    Decoding Strategy & New Estimated Hyperparameter & p-value & Perplexity Without Defense & Perplexity With Defense\\
    \midrule
    Temperature ($\tau=0.9$) & 1.0711 &  1.43e-93 & 36.659 $\pm$ 1.883 & 36.748 $\pm$ 1.827\\
    Temperature ($\tau=0.8$) & 0.9248 & 1.6312e-37 & 25.562 $\pm$ 1.573 & 26.403 $\pm$ 1.079\\
    Temperature ($\tau=0.7$) & 0.7888 & 7.135e-08 & 17.971 $\pm$ 1.213 & 19.664 $\pm$ 1.042\\
    NS ($p=0.9$) & 0.9863 & 1.3e-15 & 26.418 $\pm$ 1.164 & 28.646 $\pm$ 1.204\\
    NS ($p=0.8$) & 0.9040 & 2.9e-80 & 19.674 $\pm$ 0.812 & 21.024 $\pm$ 0.772\\
    NS ($p=0.7$) & 0.7863 & 3.2e-60 & 16.239 $\pm$ 0.842 & 17.1942 $\pm$ 0.625\\
    \bottomrule
    \end{tabular}
    }
    \end{center}
\end{table*}
\section{Ethics Discussions}

Through the course of our experiments, we did not attempt to steal the decoding algorithms of any real-world LM systems. Instead, our experiments were all performed on our own LM systems. 

Our work demonstrates the possibility of stealing IP information from real-world LM systems. 
Note that our demonstrated attacks do not target GPT-2/3/Neo,
but instead third parties who use these LMs in building their downstream tasks.
Given the abundance of such third parties, it will not be possible to contact these third parties for responsible disclosure of this vulnerability. This paper serves to disclose the threat of stealing decoding algorithms of LM-based systems to the whole  community.   
Additionally, we discussed potential  countermeasure for the API providers. 




\section{Limitations}

As previously discussed, certain stages of our attack, such as detecting greedy and beam search or identifying k in top-k sampling, do not necessitate access to inner probabilities, and these stages can be applied to any type of LMs. However, some stages do require such information, resulting in a potential limitation: the reference model used to obtain inner probabilities must be of the same type as the targeted model, as different models exhibit distinct behaviors on sequences. Most text generation APIs rely on a limited set of models, such as GPT-2, GPT-3, or GPT-Neo. Despite their distinct behaviors on identical sequences, it is possible to compare the outputs with texts generated by these models. It is important to note that decoding algorithms do not alter the order of tokens based on their probabilities; they only adjust the probabilities to assign different weights to various tokens. Consequently, detecting the base model employed by the API is not an overly challenging task.
\section{Related Works: Attacks on NLP}


Like image applications, NLP models are vulnerable to any types of attacks. Membership Inference Attacks (MIA) disclose if a data-point was used to train the victim model~\cite{shokri2017membership}. Recent works have shown how powerful these attacks are on NLP classification tasks~\cite{shejwalkar2021membership}.  Mahloujifar et al.~\cite{mahloujifar2021membership} show that word embedding is also vulnerable to MIA. Carlini et al.~\cite{carlini2021extracting} investigate memorization in large LMs that leads to data extraction attacks. Model inversion attacks~\cite{fredrikson2015model} reconstruct representative views of subset of examples.

The backdoor attack ~\cite{gu2017badnets} is another emergent issue that threats language models. In backdoor attacks, the adversary injects backdoor into language models during the training. Hence, the adversary uses a trigger in a poisoned example to activate the backdoor to make the model produces the output it wants, while the model has a normal behavior for other benign examples. Recently, many works have been done to investigate backdoor attack in language models~\cite{chen2021badpre, zhang2021trojaning, li2021backdoor, shen2021backdoor}. Besides these, two more relevant attacks to our work are model stealing/imitation and hyperparameter stealing.

\textbf{Model Stealing/Imitation/Extraction Attack.} Model stealing attacks, also called model extraction attacks or model imitation attacks, have been widely explored in simple classification~\cite{tramer2016stealing} computer vision tasks~\cite{orekondy2019knockoff}, both theoretically~\cite{milli2019model} and empirically. Model extraction attacks seek to replicate the functionality of the victim model, thereby facilitating further attacks against it. For instance, the adversary can use the extracted model to construct adversarial examples that make the model to have incorrect predictions~\cite{he2021model}.

Many works~\cite{49091, he2021model, lyu2021killing} study model extraction attacks against BERT-based APIs. Wallace et al.~\cite{wallace2020imitation} investigate model stealing attacks on machine translation by querying them in black-box setting. Most of these works solve the problem for the tasks where the output or response is more restricted or predictable, including sentiment classification, machine translation, or extractive QA tasks. Model extraction attacks has not been explored for open-ended text generation that requires the output to be more diverse. This problem is a potential research path as future works.

\textbf{Hyperparameter Stealing Attack.} Wang et al.~\cite{wang2018stealing} demonstrate that various machine learning algorithms are vulnerable to hyperparameter stealing attacks. Oh et al.~\cite{oh2018towards} propose a metamodel to infer some attributes of the Neural Network related to the architecture and training process. In LMs, hyperparameters can be classified into two groups: one comprising parameters universal to many machine learning algorithms (e.g., batch size, regularization term, k in KNN), and the other exclusive to NLP models. This paper concentrates on decoding algorithms and their  corresponding hyperparameters. Stealing hyperparameters has not been studied as much as other attacks. In a concurrent work, Ippolito et al.~\cite{ippolito2023reverse} apply a similar approach to detect the decoding algorithm if the API uses top-k or Nucleus Sampling as the decoding algorithms. However, their work primarily focuses on these two types of decoding algorithms, offering a narrower scope compared to the broader range of hyperparameters discussed in this paper. 

\vspace{-1mm}
\section{Conclusion}

In this paper, we presented the first decoding algorithm stealing attack on LMs. The decoding algorithms used in open-ended text generation are critical, and organizations are willing to invest significant resources to find the best decoding strategies and corresponding hyperparameters. This motivates adversaries to attempt to steal this information. Our results showed that it is possible to do so at a relatively low cost. We also proposed potential defense against this mathematical attack. Additionally, we highlighted that inferring certain information is possible even without direct access to the probabilities provided by the API's LM. We hope that this work brings attention to the vulnerabilities of decoding algorithms in LM-based systems and encourages further research in this area.

\section*{Acknowledgments}
This work was  supported by the
NSF grant  $2131910$.

\printbibliography

\appendix

\section{More Examples on Estimating Beam Size} \label{sec:beam_size}

In this section, we aim to provide additional examples to illustrate our approach for estimating the beam size. We hope that these examples will further clarify our algorithm. We present the prompts that result in correct estimations for beam sizes of $7$, $8$, and $9$ in Table ~\ref{tab:Beam_Size}. These prompts have been selected from the Genre-Based Story Generation Dataset and short stories generated by ChatGPT. The rank of the tokens provided at each time step can also be found in this section.

\section{Effect of the size of the model} \label{sec:size_effect}

In our research, we have found that the size of the language model does not have a significant impact on the performance of our algorithm. Our algorithm is designed to be robust to variations in model size and is able to produce consistent results regardless of the size of the language model used. Additionally, we have conducted experiments using a range of model sizes of GPT-2 and GPT-3 and have observed similar performance across all models, further supporting our claim that the size of the language model does not affect the results of our algorithm. This is an important aspect of our research as it allows for flexibility in the use of models of different sizes without compromising the performance of the algorithm. The results of some parts of our algorithm on Medium and Large GPT-2, as well as larger GPT-3 models such as Babbage and Curie are presented in Table ~\ref{tab:Size_Effect} and Table ~\ref{tab:Size_Effect2}. Please note that in the context of our research, utilizing larger variants of GPT-Neo, such as the 2.7-billion parameter model (GPT-Neo 2.7B), proves to be impractical due to the constraints posed by our limited computational resources.

\begin{figure}[t]
\centering
\includegraphics[width=8.0cm,height=\textheight,keepaspectratio]{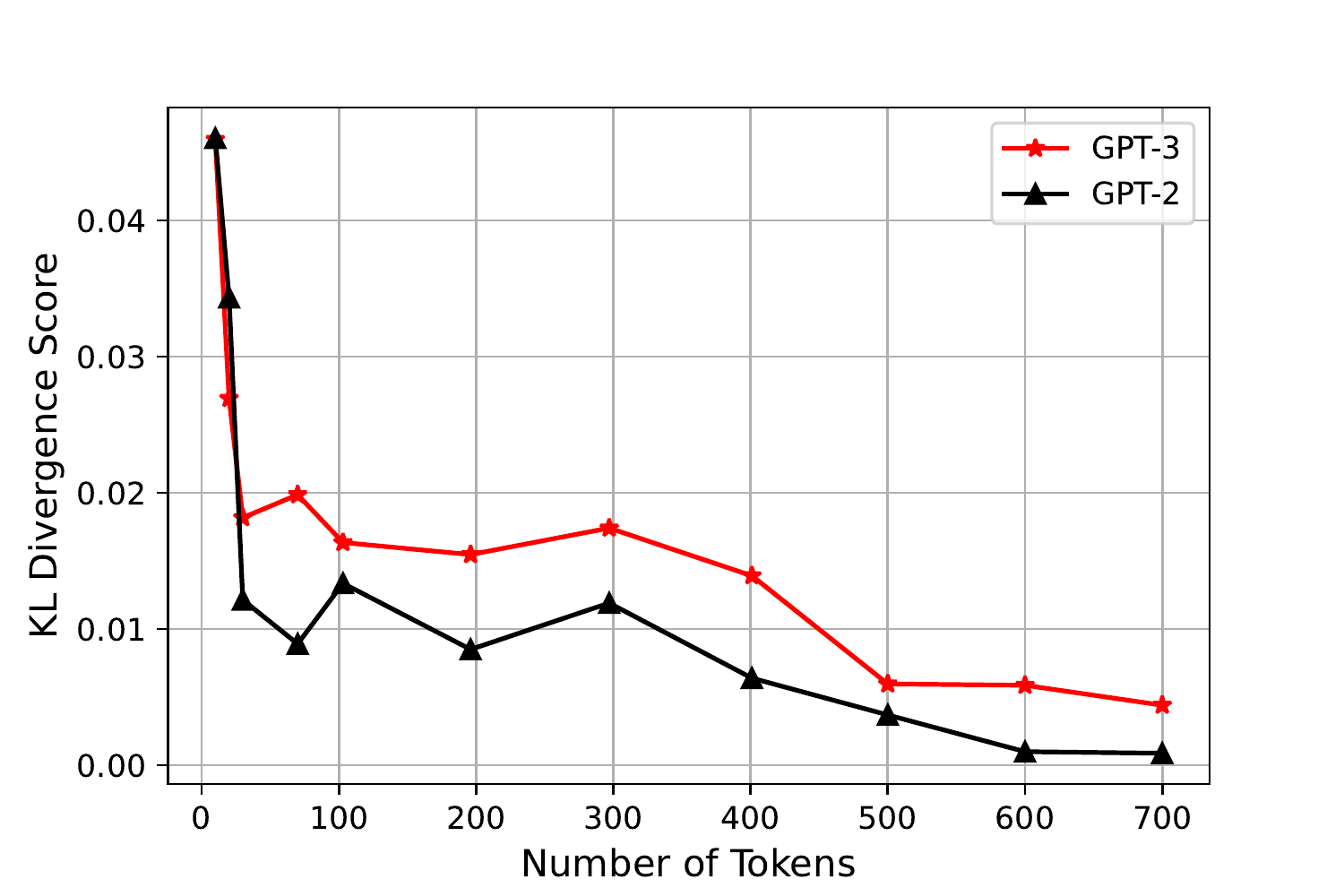}

\caption{This figure illustrates the difference in probability distributions between a victim model using prompts before a given query and a base model without prompts, as the length of the queries increases. The difference between the distributions is measured using KL-divergence and the results are presented for both GPT-2 and GPT-3 models. As can be seen in the figure, the difference between the distributions decreases with an increase in query length.}
\label{fig:ns_queries}
\end{figure}

\section{How to apply the algorithm if the API uses prompt engineering?} \label{sec:prompt_eng}

As previously discussed in Section ~\ref{sec:other_discussion} and illustrated in Figure ~\ref{fig:ns_queries}, by increasing the length of the sequence sent as a query, the probability distribution becomes more similar. Besides, Khandelwal et al.~\cite{khandelwal-etal-2018-sharp} show that long-range context does not significantly affect local decisions. We utilize this phenomenon to apply our algorithm in cases where the API uses prompt engineering to generate the desired output.
It is important to note that this is only effective when the original text and the prompt used by the API are from the same context. For example, if we use a long sequence of a drama story about a young woman and then ask a story generation API to complete it, the probability distribution will be similar. Here is the query:

\textit{"As a young girl, Sarah always felt like she was different from everyone else. She didn't quite fit in with the other kids her age, and she always had an active imagination. Little did she know, her imagination would soon become a reality. One day, while out exploring in the woods near her home, Sarah stumbled upon an old, dilapidated cottage. Curiosity getting the best of her, she decided to investigate. As she made her way inside, she was immediately struck by a feeling of power and magic. As she explored the cottage, she came across an ancient book hidden behind a loose stone in the wall. Sarah couldn't resist the urge to open it, and as she did, a bright light burst from the pages, enveloping her in its glow. When the light faded, Sarah felt a surge of energy coursing through her veins. She soon realized that she had been imbued with magical powers, and she knew that she had to use them for good. Sarah set out to learn how to control and harness her powers, and with each passing day, she grew stronger and more skilled. But she knew that she couldn't keep her powers a secret for long. One day, Sarah received a summons from the king, who had learned of her powers and needed her help. A dark sorceress was threatening to take over the kingdom, and Sarah was the only one who could stop her.  Sarah was nervous, but she knew that she couldn't let her fear get in the way of her duty. She accepted the mission and set out to confront the sorceress. As Sarah faced off against the dark sorceress, she knew that this was the moment that would define her. She summoned all of her magical energy and focused it into a single, powerful blast. The sorceress was no match for Sarah's strength, and she was defeated. The kingdom rejoiced at the news of the sorceress' defeat, and Sarah was hailed as a hero. She had saved the kingdom and proved that she was more than just a young woman with magical powers - she was a true warrior. Sarah returned home a changed person. She no longer felt like an outcast, but rather, a hero and a beacon of hope for others. And as she looked to the future, she knew that her adventures were far from over. She would continue to use her powers for good, and she would always be ready to defend her kingdom whenever it was in danger. Sarah's journey had only just begun, and she knew that there would be many challenges ahead. But she was ready to face them head on, armed with her magical powers and her unwavering determination to do good in the world. As she continued to hone her skills and learn more about her powers, Sarah became a powerful force for good in her kingdom. She used her magic to protect the innocent and bring justice to those who threatened the peace of the land. Despite the many dangers she faced, Sarah remained brave and dedicated to her cause. And as she traveled across the kingdom, she gained many loyal friends and allies who stood by her side and supported her on her journey. Through her bravery and selflessness, Sarah became a true hero and an inspiration to all those around her. She had discovered her true purpose in life and was determined to use her gifts for the greater good. And so, Sarah's adventures continued, as she worked to rid the kingdom of evil and bring about a brighter, more peaceful future for all. Sarah's adventures took her to far-off lands and through treacherous terrains, but she never lost sight of her purpose. She encountered all sorts of magical creatures and encountered powerful sorcerers and witches who sought to challenge her. But Sarah was not one to be underestimated. She had become a master of her powers, and she used them with great precision and skill. She battled fierce monsters and defeated powerful enemies, always emerging victorious. As she traveled, Sarah met many people who were in need of her help. She used her powers to heal the sick, protect the helpless, and bring hope to those who had lost it. She became known as a guardian of the innocent and a defender of justice, and her reputation grew with each passing day. Eventually, Sarah returned home to her kingdom, where she was greeted with great celebration. The people hailed her as a hero and thanked her for her bravery and selflessness. Sarah basked in the adoration of her people, but she knew that her work was far from over. She vowed to continue using her powers for good, and to always stand up for what was right, no matter the cost. And so, Sarah's "
}

It is possible that the API uses a prompt related to the story to improve the relevance of the output. An example of such a prompt added by the API could be:

\textcolor{black}{\textit{"Complete the story about a young woman who discovers she has magical powers and must learn how to use them to save her kingdom from a dark sorceress."}}

Now, we repeat some of our experiments using these queries and prompts on small GPT-2 and GPT-3 models. We query the base models using the provided text and use the generated probability distribution as the reference internal probability distribution. Then, we assume that the API adds the provided prompt to the beginning of the text to generate the desired output. The results of these experiments can be found in Table ~\ref{tab:Exp_with_Prompt}.

\section{Other Discussions}\label{sec:other_discussion2}

\textbf{Attacking a system that fine-tunes a LM without having access to the inner probability distribution.} In cases where the API fine-tunes a base model for its downstream task, and the attacker has no access to any model-stealing attacks to approximate the internal probability distribution, they can still apply some parts of our algorithm. For instance, the first and second stages can be done using only the input and the generated text. Additionally, the beam size can be extracted by analyzing the final probability distributions. Furthermore, the attacker can detect the use of top-k if it is used as the last part of the API's decoding algorithm and infer the hyperparameter k without any information about the internal probabilities.

\textbf{Decoding strategies are vulnerable?} As the number of decoding algorithms grows, our investigation focuses on a limited yet representative subset of decoding strategies. This study is a starting point for asserting that all decoding algorithms may be vulnerable. Consequently, future decoding methods must incorporate considerations for privacy and security. Furthermore, examining the vulnerabilities of state-of-the-art and novel decoding strategies~\cite{su2022contrastive, krishna2022rankgen, li2022contrastive, hewitt2022truncation} could be a valuable area for future research.

\begin{table*}[ht]
    \centering
    \caption{\label{tab:Beam_Size} An example of the results obtained when estimating the beam search method with beam sizes of $7$, $8$, and $9$. The column 'Rank' refers to the position of the highlighted token in the sorted list of tokens generated by the model. The bold rank represents the highest rank among the generated tokens, which serves as our estimation of the beam size.}
    \label{tab:Beam_Size}
    \begin{tabular}{cllc}
    \toprule
        Time Step & Generated Text  & Rank\\
        \midrule
        t & Kannathaa (Manorama), an old rich countrywoman of character, & - \\
        t+1 & Kannathaa (Manorama), an old rich countrywoman of character, \textcolor{red}{is} &  1\\
        t+2 & Kannathaa (Manorama), an old rich countrywoman of character, \textcolor{red}{has} been & 2 \\
        t+3 & Kannathaa (Manorama), an old rich countrywoman of character, \textcolor{red}{comes} to the & \textbf{7} \\
        t+4 & Kannathaa (Manorama), an old rich countrywoman of character, \textcolor{red}{comes} to the rescue & \textbf{7} \\
        t+5 & Kannathaa (Manorama), an old rich countrywoman of character, \textcolor{red}{comes} to the rescue of & \textbf{7} \\
        t+6 & Kannathaa (Manorama), an old rich countrywoman of character, \textcolor{red}{comes} to the rescue of her & \textbf{7} \\
        \midrule
        t & The film's plot revolves around Kanthaswamy (Vikram), who  & - \\
        t+1 & The film's plot revolves around Kanthaswamy (Vikram), who \textcolor{red}{is} & 1 \\
        t+2 & The film's plot revolves around Kanthaswamy (Vikram), who \textcolor{red}{is} a & 1 \\
        t+3 & The film's plot revolves around Kanthaswamy (Vikram), who \textcolor{red}{works} as a & 6 \\
        t+4 & The film's plot revolves around Kanthaswamy (Vikram), who \textcolor{red}{is} the son of & 1 \\
        t+5 & The film's plot revolves around Kanthaswamy (Vikram), who \textcolor{red}{is} the son of a & 1 \\
        t+6 & The film's plot revolves around Kanthaswamy (Vikram), who \textcolor{red}{lives} in a small village in & \textbf{8} \\
        \midrule
        t & The ship was lost, and the & - \\
        t+1 & The ship was lost, and the \textcolor{red}{crew} &  1\\
        t+2 & The ship was lost, and the \textcolor{red}{ship} was & 2 \\
        t+3 & The ship was lost, and the \textcolor{red}{crew} of the & 1 \\
        t+4 & The ship was lost, and the \textcolor{red}{rest} of the crew & \textbf{9} \\
        t+5 & The ship was lost, and the \textcolor{red}{rest} of the crew were & \textbf{9} \\
        t+6 & The ship was lost, and the \textcolor{red}{ship} was unable to return to & 2 \\
        \bottomrule
        \end{tabular}
    
\end{table*}
\begin{table*}[t]
    \begin{center}
    \caption{Results of hyperparameter estimation for Medium GPT-2 and GPT-3. This table presents the results of our estimation for the hyperparameters $\tau$ and $p$ when the API uses Medium GPT-2 or GPT-3 models.}
    \label{tab:Size_Effect}
    \begin{tabular}{ c|ccccc } 
     \toprule
    Decoding Strategy & Real Value & Estimated Value & p-value & KL Divergence & Model\\
    \midrule
    \multirow{7}{*}{Temperature} & \multirow{2}{*}{0.6} & 0.6057 $\pm$ 0.013 & 1.0 & 0.002 $\pm$ 0.001 & Medium GPT-2\\
     &  & 0.5978 $\pm$ 0.004 & 1.0 & 0.001 $\pm$ 0.007 & Medium GPT-3\\
     \cmidrule{2-6}
     & \multirow{2}{*}{0.7} & 0.7064 $\pm$ 0.019 & 0.99 $\pm$ 0.008 & 0.002 $\pm$ 0.011 & Medium GPT-2\\
     &  & 0.7024 $\pm$ 0.008  & 1.0 & 0.002 $\pm$ 0.009 & Medium GPT-3\\
     \cmidrule{2-6}
     & \multirow{2}{*}{0.85} & 0.8550 $\pm$ 0.024 & 1.0 & 0.004 $\pm$ 0.009 & Medium GPT-2\\
     &  & 0.8491 $\pm$ 0.011 & 1.0 & 0.001 $\pm$ 0.004 & Medium GPT-3\\
    \midrule
    \multirow{7}{*}{Nucleus Sampling} & \multirow{2}{*}{0.6} & 0.6027 $\pm$ 0.014 & 1.0 & 0.006 $\pm$ 0.01 & Medium GPT-2\\
     &  & 0.6026 $\pm$ 0.004 & 1.0 & 0.012 $\pm$ 0.019 & Medium GPT-3\\
     \cmidrule{2-6}
     & \multirow{2}{*}{0.8} & 0.7905 $\pm$ 0.01 & 0.97 $\pm$ 0.006 & 0.006 $\pm$ 0.012 & Medium GPT-2\\
     &  & 0.8072 $\pm$ 0.007 & 0.98 $\pm$ 0.004 & 0.01 $\pm$ 0.007 & Medium GPT-3\\
     \cmidrule{2-6}
     & \multirow{2}{*}{0.9} & 0.8922 $\pm$ 0.005 & 0.98 $\pm$ 0.003 & 0.004 $\pm$ 0.002 & Medium GPT-2\\
     &  & 0.9020 $\pm$ 0.006 & 1.0 & 0.004 $\pm$ 0.01 & Medium GPT-3\\
    \bottomrule
    \end{tabular}
    \end{center}
    
\end{table*}
\begin{table*}[t]
    \begin{center}
    \caption{Results of hyperparameter estimation for Large GPT-2 and GPT-3. This table presents the results of our estimation for the hyperparameters $\tau$ and $p$ when the API uses Large GPT-2 or GPT-3 models.}
    \label{tab:Size_Effect2}
    \begin{tabular}{ c|ccccc } 
     \toprule
    Decoding Strategy & Real Value & Estimated Value & p-value & KL Divergence & Model\\
    \midrule
    \multirow{7}{*}{Temperature} & \multirow{2}{*}{0.6} & 0.6039 $\pm$ 0.005 & 1.0 & 0.001 $\pm$ 0.012 & Large GPT-2\\
     &  & 0.5951 $\pm$ 0.007 & 0.99 $\pm$ 0.012 & 0.003 $\pm$ 0.009 & Large GPT-3\\
     \cmidrule{2-6}
     & \multirow{2}{*}{0.7} & 0.7036 $\pm$ 0.006 & 1.0 & 0.007 $\pm$ 0.011 & Large GPT-2\\
     &  & 0.6961 $\pm$ 0.01 & 1.0 & 0.007 $\pm$ 0.021 & Large GPT-3\\
     \cmidrule{2-6}
     & \multirow{2}{*}{0.85} & 0.8402 $\pm$ 0.008 & 0.97 $\pm$ 0.01 & 0.01 $\pm$ 0.021 & Large GPT-2\\
     &  & 0.8446 $\pm$ 0.003 & 0.98 $\pm$ 0.013 & 0.008 $\pm$ 0.017 & Large GPT-3\\
    \midrule
    \multirow{7}{*}{Nucleus Sampling} & \multirow{2}{*}{0.6} & 0.593 $\pm$ 0.009 & 0.99 $\pm$ 0.006 & 0.009 $\pm$ 0.013 & Large GPT-2\\
     &  & 0.6047 $\pm$ 0.005 & 1.0 & 0.001 $\pm$ 0.009 & Large GPT-3\\
     \cmidrule{2-6}
     & \multirow{2}{*}{0.8} & 0.7936 $\pm$ 0.003 & 0.99 $\pm$ 0.002 & 0.001 $\pm$ 0.007 & Large GPT-2\\
     &  & 0.7949 $\pm$ 0.005 & 0.99 $\pm$ 0.002 & 0.009 $\pm$ 0.016 & Large GPT-3\\
     \cmidrule{2-6}
     & \multirow{2}{*}{0.9} & 0.8932 $\pm$ 0.004 & 0.98 $\pm$ 0.011 & 0.013 $\pm$ 0.025 & Large GPT-2\\
     &  & 0.8938 $\pm$ 0.007 & 0.98 $\pm$ 0.009 & 0.002 $\pm$ 0.011 & Large GPT-3\\
    \bottomrule
    \end{tabular}
    \end{center}
    
\end{table*}

\begin{table*}[t]
    \begin{center}
    \caption{This table presents the results of our estimation for the hyperparameters $p$ and $\tau$ when the API uses prompt engineering and the attacker does not have access to the internal probability distribution. The results are for both GPT-2 and GPT-3 models.}
    \label{tab:Exp_with_Prompt}
    \begin{tabular}{ c|ccccc } 
     \toprule
    Decoding Strategy & Real Value & Estimated Value & p-value & KL Divergence & Model\\
    \midrule
    \multirow{3}{*}{Temperature} & 0.65 & 0.6402 $\pm$ 0.008 & 0.98 $\pm$ 0.012 & 0.007 $\pm$ 0.003 & GPT-2\\
    \cmidrule{2-6}
     & 0.8  & 0.7933 $\pm$ 0.011 & 0.99 $\pm$ 0.011 & 0.011 $\pm$ 0.008 & GPT-2\\
     \cmidrule{2-6}
     & 0.9 & 0.8916 $\pm$ 0.011 & 0.97 $\pm$ 0.018 & 0.023 $\pm$ 0.011 & GPT-2\\
    \midrule
    \multirow{7}{*}{Nucleus Sampling} & \multirow{2}{*}{0.7} & 0.7245 $\pm$ 0.006 & 0.97 $\pm$ 0.023 & 0.011 $\pm$ 0.005 & GPT-2\\
     &  & 0.7116 $\pm$ 0.037 & 0.98 $\pm$ 0.013 & 0.009 $\pm$ 0.002 & GPT-3\\
     \cmidrule{2-6}
     & \multirow{2}{*}{0.8} & 0.8206 $\pm$ 0.005 & 0.96 $\pm$ 0.027 & 0.011 $\pm$ 0.009 & GPT-2\\
     &  & 0.7983 $\pm$ 0.017 & 0.99 $\pm$ 0.011 & 0.007 $\pm$ 0.002 & GPT-3\\
     \cmidrule{2-6}
     & \multirow{2}{*}{0.9} & 0.9163 $\pm$ 0.005 & 0.97 $\pm$ 0.009 & 0.021 $\pm$ 0.011 & GPT-2\\
     &  & 0.8905 $\pm$ 0.018 & 0.98 $\pm$ 0.009 & 0.01 $\pm$ 0.006 & GPT-3\\
    \bottomrule
    \end{tabular}
    \end{center}
    
\end{table*}

\end{document}